\documentclass[11pt,a4paper]{article}
\usepackage[hyperref]{emnlp2020}
\usepackage{times}
\usepackage{latexsym}

\usepackage{microtype}

\aclfinalcopy % Uncomment this line for the final submission

\def\aclpaperid{2164} %  Enter the acl Paper ID here

\usepackage{graphicx}
\usepackage{algorithm}
\usepackage{algorithmicx}
\usepackage{algpseudocode}
\usepackage{multirow}
\usepackage{xcolor}
\usepackage{tikz}
\usepackage{amsmath,amssymb,mathtools}
\usetikzlibrary{positioning,calc,shapes.geometric,backgrounds,arrows.meta}
\usepackage{bm}
\usepackage{CJKutf8}
\usepackage{comment}
\usepackage{booktabs,multirow}
\usepackage{amsthm}
\usepackage[normalem]{ulem}

\usepackage{array}
\usepackage{colortbl}
\usepackage[whole]{bxcjkjatype}
\usepackage{pifont}
\usepackage{setspace}
\usepackage{enumitem}
\setlist{itemsep=1pt, topsep=3pt}
\setlist[itemize]{leftmargin=*}
\setlist[enumerate]{leftmargin=*}

\usepackage{notation}

\newcounter{magicrownumbers}
\newcommand\rownumber{\scriptsize{\stepcounter{magicrownumbers}\arabic{magicrownumbers} :}}
\newcommand\setcount{\setcounter{magicrownumbers}{0}}

\newcommand{\cmark}{\ding{52}}
\newcommand{\xmark}{\ding{56}}

\newcolumntype{+}{>{\global\let\currentrowstyle\relax}}
\newcolumntype{^}{>{\currentrowstyle}}

\newcolumntype{H}{@{}>{\lrbox0}l<{\endlrbox}}

\title{Filtering Noisy Dialogue Corpora\\by Connectivity and Content Relatedness}

\author{
    Reina\,Akama$^{1,2}$ \quad
    Sho\,Yokoi$^{1,2}$ \quad
    Jun\,Suzuki$^{1,2}$ \quad
    Kentaro\,Inui$^{1,2}$
    \\
    $^{1}$Tohoku University \quad
    $^{2}$RIKEN \quad
    \\
    \texttt{\{reina.a, yokoi, jun.suzuki, inui\}@ecei.tohoku.ac.jp} 
    \\
}

\date{}

\begin{document}
\maketitle

\begin{abstract}
Large-scale dialogue datasets have recently become available for training neural dialogue agents. 
However, these datasets have been reported to contain a non-negligible number of unacceptable utterance pairs. 
In this paper, we propose a method for scoring the quality of utterance pairs in terms of their connectivity and relatedness. 
The proposed scoring method is designed based on findings widely shared in the dialogue and linguistics research communities. 
We demonstrate that it has a relatively good correlation with the human judgment of dialogue quality. 
Furthermore, the method is applied to filter out potentially unacceptable utterance pairs from a large-scale \emph{noisy} dialogue corpus to ensure its quality. 
We experimentally confirm that training data filtered by the proposed method improves the quality of neural dialogue agents in response generation.%
    \footnote{The code is publicly available at \url{https://github.com/jqk09a/CoRe-dialogue-filtering}.}
\end{abstract}

%%%%%%%%%%%%%%%%%%%%%%%%%%%%%%%%%%
\section{Introduction}
\label{sec:introduction}
%%%%%%%%%%%%%%%%%%%%%%%%%%%%%%%%%%%%

Some million-scale datasets such as movie scripts and social media posts have become available in recent years for building neural dialogue agents~\cite{lison2016lrec:opensubtitles,henderson2019convai:convdata}. 
Such large-scale datasets can be expected to improve the performance of dialogue response generation models based on deep neural networks (DNNs) since the combination of DNNs and large-scale training datasets has led to considerable performance improvement in many sentence generation tasks~\citet{koehn2017nmt:sixchallenges,sennrich2019acl:nmtcasestudy,adiwardana2020arxiv:meena}.

\begin{figure}[t]
	\centering
	\includegraphics[width=0.95\linewidth]{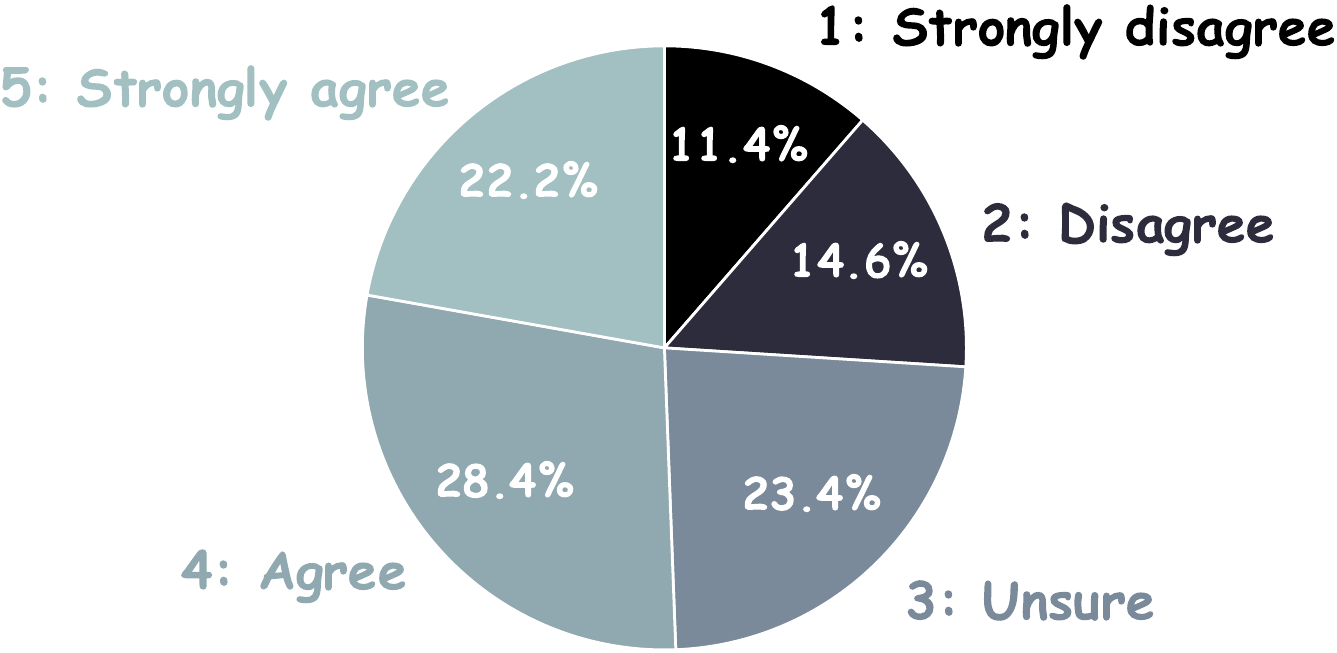}
    \vskip -0.5mm
    \caption{\textit{Is the sequence of the two utterances acceptable as a dialogue?} Response acceptability scores are given by humans on the English OpenSubtitles corpus.}
    \label{fig:preliminary experiment}
    \vskip -2mm
\end{figure}

In contrast to the quantity of the data, the quality of the data has often been problematic.
For example, OpenSubtitles~\cite{lison2016lrec:opensubtitles,lison2018lrec:opensubtitles}, the most widely used large-scale English dialogue corpus, was constructed by collecting two consecutive lines of movie subtitles under the simplified assumption that one line of a movie subtitle is one utterance and the next line is the next utterance follow it.
Inevitably, this corpus includes unacceptable utterance pairs from the viewpoint of a conversational sequence, e.g., caused by scene switching or flashback.
Several previous studies have identified such flaws and reported that the corpus is \emph{noisy}~\cite{vinyals2015icml:neuralconv,li2016naacl:diversity,baheti2018emnlp:generating-interesting-response}, where \emph{noisy} refers to unacceptable utterance pairs in this context.
Figure~\ref{fig:preliminary experiment} shows the result of our experimental investigation regarding the acceptability rate of the utterance pairs in the OpenSubtitles corpus.%
    \footnote{See Appendix~\ref{a:preliminary experiment} for detailed experimental settings.}
It can be noticed from the figure that only half of the utterance pairs can be considered \textit{acceptable} (i.e., were rated with score $5$: Strongly agree or $4$: Agree), and over 25\% of utterance pairs are clearly \textit{unacceptable} (i.e., were rated with score $1$: Strongly disagree or $2$: Disagree) from the human perspective.%
    \footnote{See Table~\ref{tab:preliminary scored pairs} for samples of unacceptable/acceptable utterance pairs annotated by humans.}

With this situation, a straightforward research question arises, namely, \emph{Can we further improve the performance of neural response generation models by ablating unacceptable utterance pairs from training data?}
To the best of our knowledge, no previous study has explicitly focused on this question.
Thus, the goal of this paper is to provide an answer to this question.
Furthermore, it is not clear whether and how one can effectively discover unacceptable utterance pairs within large-scale training datasets.
This study explores a way of constructing a scoring method for filtering \emph{noisy} data filtering to improve the performance of response generation models.

To achieve the set goals, we started with a review of previous arguments about the criteria for identifying appropriate utterances in dialogues and designed our scoring function that is consistent with reflects as much of the community's consensus as possible.
In particular, the proposed scoring method estimates the quality of utterance pairs based on the following two aspects: 
(i) the \textbf{connectivity} between source and target utterances and 
(ii) their \textbf{content relatedness} (Section~\ref{sec:idea}).

The contributions of this study are the following:
\begin{itemize}
    \item We propose a scoring method for estimating the quality of utterance pairs in an unsupervised manner (Section~\ref{sec:proposed method});
    \item We reveal that our scoring method effectively detects unacceptable utterance pairs, and thus, be appropriate for noisy data filtering (Section~\ref{sec:experiments});
    \item We empirically prove that our proposed data filtering method improves the performance of neural response generation models (Section~\ref{sec:case study}); and
    \item We confirm that our noisy data filtering approach is effective across different languages and dataset sizes (Section~\ref{sec:Japanese}).
\end{itemize}

%%%%%%%%%%%%%%%%%%%%%%%%%%%%%%%%%%%%%
\section{Task Definition: Noisy Data Filtering}
\label{sec:task}
%%%%%%%%%%%%%%%%%%%%%%%%%%%%%%%%%%%%%

Let $x$ be an \textbf{utterance} and $y$ be a \textbf{response} to $x$.
Then, an \textbf{utterance pair} can be denoted as we refer to $(x,y)$.
Let $\mathcal{D}$ be a dataset that comprising a set of {utterance pairs}, $\mathcal D = \{(x,y)\}$.
Then, the task can be formulated as ablating unacceptable utterance pairs from $\mathcal D$ to obtain a less noisy subset $\mathcal{D}'\subseteq \mathcal{D}$, hereinafter referred to as filtered dataset.
$\mathcal{D}'$ can then be used to train response generation models. 
This paper refers to this process as \textbf{noisy data filtering}, where \emph{noisy} means unacceptable utterance pairs in this context. 
Furthermore, we establish a function $S\colon \mathcal D\to \mathbb R$ is used to score the degree of \textit{acceptability} of each utterance pair $(x,y) \in \mathcal D$.

%%%%%%%%%%%%%%%%%%%%%%%%%%%%%%%%%%%%
\section{Background}
\label{sec:background}
%%%%%%%%%%%%%%%%%%%%%%%%%%%%%%%%%%%%

\paragraph{Response generation using noisy data.}
The following two approaches are widely used to address the problem of dialogue response generation noisy dialogue corpora. 
According to the \emph{model approach}, models are trained while handling noise at the same time. 
For example, \citet{shang2018ijcai:calibration} proposed a method with a calibration framework and demonstrated its effectiveness on a Chinese corpus.
According to the \emph{data approach}, training data are pre-processed with the aim of improving their quality before training models.
In this study, we take the data approach in light of the success of noisy parallel corpus filtering in machine translation (MT). 
Additionally, it has become a reasonable strategy to reduce the size of training data since enormous dialogue data has been available.
\citet{csaky2019acl:filtering}'s method is most relevant to our study in that it cleanses dialogue corpora.
However, the main goal of their method is to eliminate generic, or boring, responses, whereas the goal of the method proposed here is to eliminate unacceptable utterance pairs. 
This difference in goals leads to the essential difference in filtering strategies.

\paragraph{Effectiveness of filtering noisy data in neural machine translation.}

Researchers in the field of neural machine translation (NMT) have recognized that collecting high-quality training data to be equally or even more important than exploring sophisticated model architectures~\cite{koehn2018wmt:filteringfindings,junczys-dowmunt2018wmt:filtering, morishita2018wmt:ntt}.
Techniques used in neural response generation and NMT are nearly identical; e.g., sequence-to-sequence models~\cite{Sutskever2014nips:seq2seq} and Transformers~\cite{vaswani2017nips:transformer} are often used as base model architectures.
We hypothesize that high-quality filtered dialogue data can also improve the performance of dialogue response generators.
However, the straightforward application of methods proposed for filtering noisy data in NMT may not work well due to the different nature of NMT and neural response generation tasks.
In particular,  MT data have one-to-one (ignoring paraphrases) correspondence in source and target sentences, whereas dialogues have many-to-many mappings~\cite{zhao2017acl:onetomany}.
The experiments presented in this paper provide an answer to whether NMT filtering methods can perform well in dialogue response generation.

%%%%%%%%%%%%%%%%%%%%%%%%%%%%%%%%%%%%
\section{Requirements to Utterance Pairs}
\label{sec:idea}
%%%%%%%%%%%%%%%%%%%%%%%%%%%%%%%%%%%%

In this section, we investigate the requirements that should be satisfied by an acceptable utterance pair.

\newcommand{\bgf}[1]{\tikz[baseline=(X.base)]{\node(X)[rectangle, fill=cyan!60!blue!12, rounded corners, text height=0.8ex]{#1};}}
\newcommand{\topic}[1]{{\scriptsize \texttt{[#1]}}}

\begin{table*}[!t]
    \centering
    \small
    \renewcommand{\arraystretch}{1.3}
    \tabcolsep 4pt
        \setcount
        \begin{tabular}{rlp{54ex}c}
            \toprule
            & Utterance & Response & Human \\
            \midrule
            \rownumber 
            & It'll be like you never left. \topic{??}
            & I painted a white line on the street way over there.~\topic{painting}
            & 1.4 \\

            \rowcolor{gray!7}
            \rownumber
            & You're gonna get us assimilated. \topic{??}
            & Switch to a garlic shampoo. \topic{??}
            & 1.8 \\

            \midrule

            \rownumber 
            & I probably asked for too much money. \topic{money}
            & Money's always a problem, isn't it? \topic{money}
            & 4.2 \\

            \rowcolor{gray!7}
            \rownumber
            & \bgf{I wonder} who \bgf{I should} call back. \topic{phone}
            & \bgf{They're saying they want to} call one of you back. \topic{phone}
            & 4.4 \\

            \rownumber 
            & Okay, so \bgf{where's} the rest? \topic{??}
            & Electronically scanned and archived \bgf{at headquarters} but you'll have to speak with them about that. \topic{work}
            & 4.4 \\

            \bottomrule
        \end{tabular}
    \caption{Samples of pairs judged as unacceptable/acceptable in preliminary experiments. Human denotes the average score of five human evaluators on a scale of $1$-$5$. Phrases considered to contribute to connectivity are \bgf{highlighted}. Estimated \topic{topic} of utterance is written in the end of each utterance.}
    \label{tab:preliminary scored pairs}
\end{table*}

% ==================================
\subsection{Criteria for Manual Evaluation}
\label{ssec:criteria for manual evaluation}

The instructions for manual evaluation provided by the dialogue community explain the key factors for distinguishing acceptable and unacceptable utterance pairs.

In many previous studies, human raters were asked to evaluate the \textbf{connectivity} of utterance pairs.
For instance, \citet{shang2015aclijcnlp:neuralresponding} asked whether a response could be considered as \textit{an appropriate and natural response to the post}.
\citet{xing2017aaai:topicaware} asked whether \textit{the response can be used as a reply}.
\citet{pei2018emnlp:s2spmn} asked whether \textit{the answer is natural} for the question.
Other studies have also evaluated the same or similar aspects by using keywords related to the connectivity, such as \textit{semantically appropriate for}~\cite{akama2017ijcnlp:generating} or \textit{coherent with}~\cite{shen2017acl:conditional-variational}, and \textit{coherence}~\cite{lowe2017acl:autoturingtest}.

Another frequently used metric is \textbf{content relatedness}.
For instance, \citet{galley2015aclijcnlp:deltableu} asked human evaluators to evaluate \textit{each response in terms of their relevance to a given utterance}.
\citet{li2016naacl:diversity} asked for the preference of responses \textit{that were more specific to certain utterances}. 
\citet{ritter2011:datadriven} suggested that \textit{an appropriate response should be on the same topic as the utterances}.
Several other studies have also focused on evaluating the \textit{relevance} between an utterance and its response~\cite{xu2018naacl:lsdscc,pei2018emnlp:s2spmn,lowe2017acl:autoturingtest}.

In summary, the most widely used criteria can be categorized into connectivity and content relatedness of utterance pairs. 
In fact, these two aspects are considered in the field of sociolinguistics as crucial features of conversation~\cite{sacks1989humanstudies:conversationalrule,sidnell2010book:conversation}.

% ==================================
\subsection{Observation}
\label{ssec:observation}

Furthermore, we investigated how the two aforementioned aspects can be observed in actual utterance pairs.
For this investigation, we use the utterance pairs scored by human raters that were used in our preliminary experiments shown in Figure~\ref{fig:preliminary experiment}.
Some examples are shown in Table~\ref{tab:preliminary scored pairs}.

We observe that typical phrase pair patterns can often be found in utterance pairs with high scores.
For example, the pair (\bgf{\textit{where is}}, \bgf{\textit{at}}) in Table~\ref{tab:preliminary scored pairs} is one of the typical phrase pair patterns that asks a place and provides an answer to it.
Other typical examples include (\textit{why}, \textit{because}) and (\textit{what do you want}, \textit{I want}).
In discourse linguistics, such phrase pair patterns are known as the concept of \textit{cohesive devices}.
Hereafter, we refer to such a typical phrase pair pattern as \textbf{key phrase pair}.

Moreover, in high scored utterance pairs, both an utterance and response are on the \textbf{same topic}.
For example, in the third example listed in Table~\ref{tab:preliminary scored pairs}, both the utterance and response mention \texttt{[money]}.

%%%%%%%%%%%%%%%%%%%%%%%%%%%%%%%%%%%%
\section{Proposed Method}
\label{sec:proposed method}
%%%%%%%%%%%%%%%%%%%%%%%%%%%%%%%%%%%%

As per the discussion in the previous section, each acceptable utterance pair should satisfy the following criteria:
\begin{itemize}
    \item \textbf{connectivity} --- existence of key phrase pairs
    \item \textbf{content relatedness} --- topic commonality
\end{itemize}
This section presents the proposed scoring functions to assess the degree of satisfying the above two criteria in an unsupervised manner.%
\footnote{The reason for focusing on an unsupervised approach the lack of data that can provide good supervision for utterance pair evaluation.}

% ==================================
\subsection{Connectivity}
\label{sec:connectivity}
Let $f$ and $e$ represent phrases obtained from $x$ and $y$, respectively.
Let $\phi(x, y)$ be a function that returns a set of all possible phrase ($n$-gram) pairs obtained from the utterance pair $(x, y)$.
We can define a finite set of all possible phrase pairs obtained from the entire dialogue data $\mathcal{D}$ as $\overline{\mathcal{P}}^{}_{\mathcal{D}} =\bigcup_{(x,y)\in\mathcal{D}} \phi(x, y)$.
Then, let $\mathcal{P}$ represent a set of key phrase pairs (defined in Section \ref{ssec:observation}).
We assume that $\mathcal{P}$ is a subset of $\overline{\mathcal{P}}_{\mathcal{D}}$, i.e., $\mathcal{P}\subseteq \overline{\mathcal{P}}_{\mathcal{D}}$.

To obtain $\mathcal{P}$, we take advantage of a phrase table extraction technique developed in statistical machine translation, e.g., Moses~\cite{koehen2017acl:moses}.
In this task, we require only some phrase pairs that can contribute to the connectivity of an utterance pair (as mentioned in Section \ref{ssec:observation}), unlike the translation task where the whole sentence must correspond in mutual.
Accordingly, in our experiments, we set the null alignment ratio (i.e., probability of no alignment) to $0.5$ and extend the phrase extraction algorithm to include only the explicitly corresponding range as phrases in our table.

Then, we define the scoring function $\Sframe$ to estimate connectivity as: 
\newcommand{\LEN}[1]{\lvert{#1}\rvert}
\begin{align}
\label{eq:s_frame}
    \Sframe(x,y) := 
    \sum_{\mathclap{(f,e) \in \phi(x, y)\cap\mathcal{P} }} 
    \hspace{2pt} 
        \max\bigl(\mathrm{nPMI}(f,e),0\bigr) \cdot \frac{\LEN{f}}{\LEN{x}} \cdot \frac{\LEN{e}}{\LEN{y}}
    \text{,}
\end{align}
where $\LEN{\cdot}$ denotes the number of words in the phrase or utterance.
To calculate the co-occurrence, we use the normalized pointwise mutual information (nPMI)~\cite{bouma2009gscl:npmi}, which normalizes the value so that low-frequency phrases do not take an extremely large value.
Note that we ignore the negative nPMI scores by the $\max(\cdot, 0)$ operation because we aim only to consider the positive effect of connectivity. 
The intuition behind Equation~\ref{eq:s_frame} is as follows:
\begin{itemize}
    \setlength{\itemindent}{0mm}
	\setlength{\parskip}{0.3mm}
    \item If a phrase pair $(f,e)$ has a high co-occurrence, the association strength of $(x,y)$ including $(f,e)$ might also be high. 
    \item If a phrase $f$ or $e$ occupies almost the entire sentence $x$ or $y$, $(f,e)$ is a strong indicator of the association of $(x, y)$.
\end{itemize}

% ==================================
\subsection{Content Relatedness}
Let $\VEC v(x)$ and $\VEC v(y)$ be sentence vector of $x$ and $y$, respectively.
We compute topic commonality of $x$ and $y$, that is, content relatedness as follows:
\begin{align}
\label{eq:s_content}
    \Scontent(x,y) := 
    \max\bigl(\cos(\VEC v(x), \VEC v(y)), 0\bigr)
    \text{.}
\end{align}
Cosine similarity between certain kinds of sentence vectors is known to be a good proxy of the topical relatedness of two sentences~\cite{Conneau2017emnlp:universalrepresentation,subramanian2018iclr:generalpurpose,xu2018emnlp:filtering}.
For the same reasons as Equation~\ref{eq:s_frame}, we ignore the negative $\cos$ scores by the $\max(\cdot, 0)$ operation.

% ==================================
\subsection{Summary}

Eventually, combining the above two scoring measures, we propose the following function:
\begin{align}
\label{eq:score}
    \Sours(x,y) := \alpha \Sframe(x,y) + \beta \Scontent(x,y)
    \text{,}
\end{align}
where $\alpha,\,\beta\in\mathbb R_{\geq 0}$ are hyperparameters that weigh the two viewpoints. 
For our experiments, we fix $\alpha$ and $\beta$ as follows:
\begin{align}
    \label{eq:score hyp a}
    & \alpha \!=\! \frac 1 {\frac 1 {\lvert\mathcal D\rvert} \! \displaystyle \sum_{\mathclap{(x,y)\in\mathcal D}} \Sframe(x,y)\!},\;
    \beta \!=\! \frac 1 {\frac 1 {\lvert\mathcal D\rvert} \! \displaystyle \sum_{\mathclap{(x,y)\in\mathcal D}} \Scontent(x,y)}
    \text{.}\!
\end{align}

%%%%%%%%%%%%%%%%%%%%%%%%%%%%%%%%%%%%
\section{Experiments: Data Scoring}
\label{sec:experiments}
%%%%%%%%%%%%%%%%%%%%%%%%%%%%%%%%%%%%

In this section, we describe our experiments that validate the effectiveness of the proposed scoring method.

\subsection{Experimental Setup}

% ==================================
\paragraph{Dataset.}
\label{ssec:dataset}

We conducted our experiments on a noisy English dialogue corpus from OpenSubtitles~\cite{lison2018lrec:opensubtitles} containing roughly $441$M lines.
As explained in Section~\ref{sec:introduction}, this corpus includes many unacceptable utterance pairs (Section~\ref{sec:introduction}).
We first applied several rule-based filtering as rudimentary preprocesses, which are typically used in the related literature.
Then, we obtained $79,\!445,\!453$ utterance pairs as our training data, which excludes our test and validation data.% 
    \footnote{See Appendix~\ref{a:corpus_creation} for details on our data such as the preparation procedure and statistics.}

\begin{figure*}[!ht]
    \centering
    \small
    \begin{tabular}{c}
        \begin{minipage}{0.33\hsize}
            \centering
            \includegraphics[trim = 0 3mm 0 0, width=0.97\columnwidth]{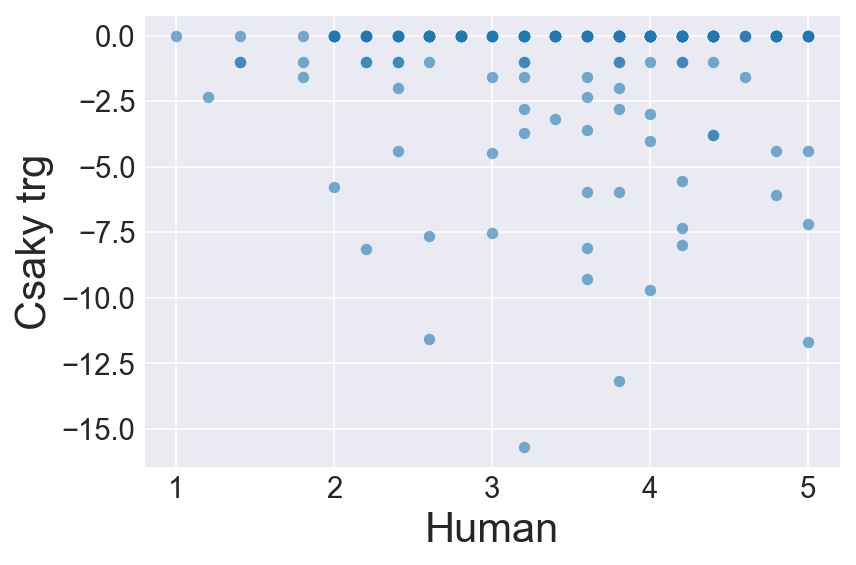}
            \hspace{20mm} (a) \quad \citet{csaky2019acl:filtering} TRG
        \end{minipage}
        \begin{minipage}{0.33\hsize}
            \centering
            \includegraphics[trim = 0 3mm 0 0, width=0.97\columnwidth]{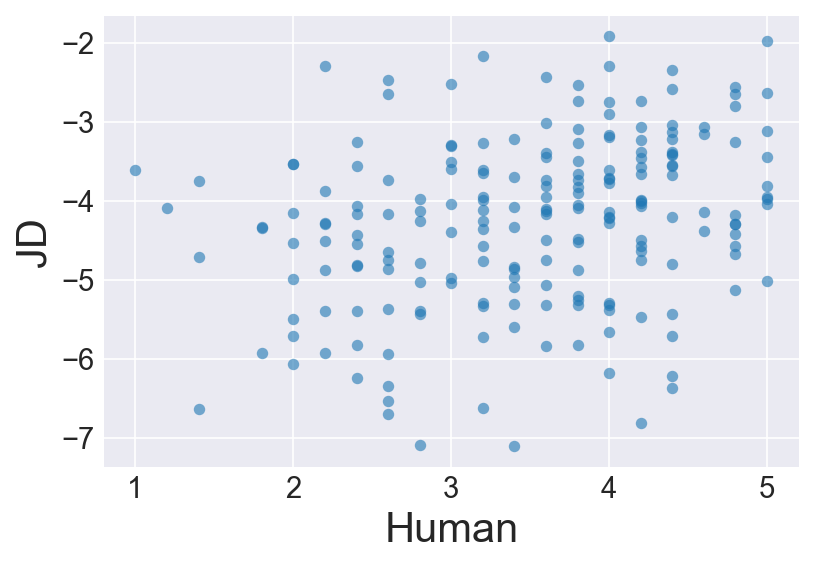}
            \hspace{20mm} (b) \quad \citet{junczys-dowmunt2018wmt:filtering}
        \end{minipage}
        \begin{minipage}{0.33\hsize}
            \centering
            \includegraphics[trim = 0 3mm 0 0, width=0.97\columnwidth]{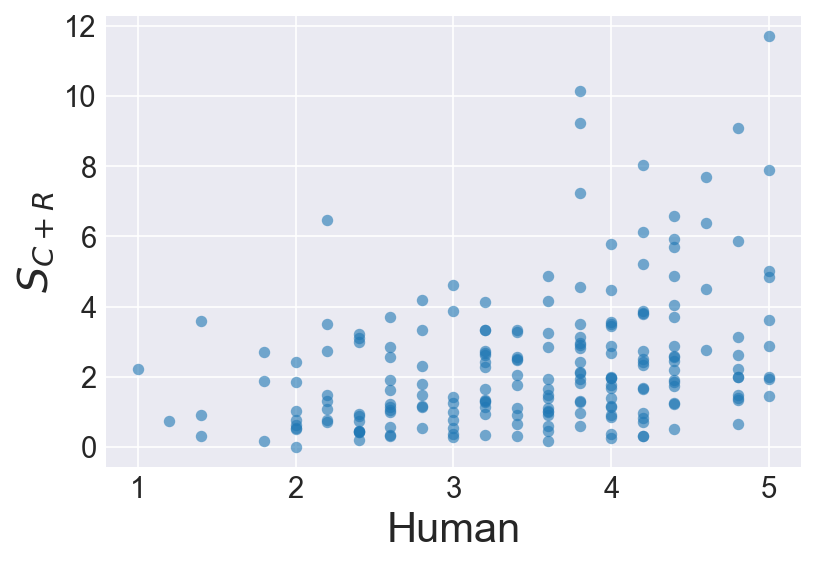}
            \hspace{20mm}  (c) \quad Ours $\Sours$
        \end{minipage}
    \end{tabular}
    \vskip -1.5mm
    \caption{Distributions between human scores and automatically computed scores by each method (English).}
    \label{fig:en_correlation-human-3}
    \vskip -0.5mm
\end{figure*}

% ==================================
\paragraph{Proposed method: detailed setup.}

To compute the connectivity $\Sframe$, we obtained a phrase table on our training data by using Moses~\cite{koehen2017acl:moses} with fastAlign~\cite{dyer2013naacl:fastalign}. 
We then removed phrase pairs with a low co-occurrence frequency (here, less than 200 times) or composed of the same phrases from the table.
As a result, the phrase table included $68,\!891$ phrase pairs, which were used as the key phrase set $\mathcal{P}$ as described in Section~\ref{sec:connectivity}.

To compute the content relatedness $\Scontent$, we created a sentence vector from pre-trained fastText word embeddings~\cite{Bojanowski2017tacl:fasttext,mikolov2018lrec:advances-in-wordrep} following \citet{arora2017iclr:sif}'s method, i.e., using SIF weighting and common component removal.
Their method is reported to be useful for computing the relatedness of two given sentences and used in many studies
~\cite{marelli2014lrec:sick,marelli2014semeval:compositional-distributional-model,Conneau2017emnlp:universalrepresentation,subramanian2018iclr:generalpurpose, baheti2018emnlp:generating-interesting-response}.
We learned common components using $30$K sentences randomly selected from the training costs appropriately. 
We then removed the first common component for all sentence vectors.

% ==================================
\paragraph{Baselines.}

For comparison, we prepared the following:
\begin{itemize}
    \item \citet{csaky2019acl:filtering}: Entropy-based filtering to remove generic utterances from the training data for promoting less-boring response generation. SRC/TRG indicates that using the entropy of source/target utterances.
    \item \citet{junczys-dowmunt2018wmt:filtering}: Filtering for NMT based on the dual conditional cross-entropy computed by a neural encoder-decoder model. It achieved the best performance on the Parallel Corpus Filtering Task at WMT~2018.%
        \footnote{\url{http://www.statmt.org/wmt18/}}
\end{itemize}

% ==================================
\paragraph{Human evaluation.}
\label{ssec:scoring eval}

To validate the ability of the proposed method to estimate the quality of utterance pairs, we measured the correlation between its scores and those assigned by humans through crowdsourcing.
We used Amazon Mechanical Turk.%
    \footnote{\url{https://www.mturk.com/}}
We randomly extracted $200$%
    \footnote{Same size as \citet{sedoc2019naacldemo:chateval,cho2020acl:yesand}.}
scored utterance pairs and asked native English-speaking crowdworkers to answer the following question for each pair: \textit{Is the sequence of the two utterances acceptable as a dialogue?} 
Workers were instructed to provide an answer on a five-point Likert scale (from $5$: Strongly agree to $1$: Strongly disagree)~\cite{likert1932archivesofpsycho:scale}.
Unqualified workers were filtered out using attention checks.
Eventually, we used the average of the scores provided by five workers as the human score for each pair.

\begin{table}[t]
    \centering
    \small
    \begin{tabular}{lrc}
        \toprule
        Scoring method & Spearman's $\rho$ & $p$-value \\
        \midrule
        \citet{csaky2019acl:filtering} SRC
        & $-$0.1173 
        & $9.8 \times 10^{-2}$ \\ 
        \citet{csaky2019acl:filtering} TRG
        & 0.0462 
        & $5.2 \times 10^{-1}$ \\ 
        \citet{junczys-dowmunt2018wmt:filtering}
        & 0.2973 
        & $1.9 \times 10^{-5}$  \\ 
        Ours $\Sours$
        & \textbf{0.3751} 
        & $\mathbf{4.4 \times 10^{-8}}$ \\ 
        
        \midrule
        
        Ours $\Sframe$ \scriptsize{(ablation study)}
        & 0.2044 
        & $3.7 \times 10^{-3}$ \\ 
        Ours $\Scontent$ \scriptsize{(ablation study)}
        & 0.3007 
        & $1.5 \times 10^{-5}$ \\ 
        \bottomrule
    \end{tabular}
    \caption{Correlation coefficient between human scores and automatically computed scores (English).}
    \label{tab:scored correlation}
\end{table}

\begin{table*}[!t]
    \centering
    \small
    \tabcolsep 3.5pt
        \setcount
        \begin{tabular}{rp{35ex}p{38ex} l rrr l c}
            \toprule
            & Utterance & Response && $\Sframe$\, & $\Scontent$ & $\Sours$ && Human \\
            \cmidrule{1-3} \cmidrule{5-7} \cmidrule{9-9}
            
            \rownumber \rule{0pt}{3ex}
            & What is the anarchy facing the jail of the sick passion? 
            & Gosh, it's really cold!
            && 0.32 & 0.00 & \textbf{0.32} && 1.4  \\
            
            \rowcolor{gray!7}
            \rownumber \rule{0pt}{3ex}
            & Pushers won't let the junkie go free. 
            & Across 110th Street.
            && 0.00 & 0.42 & \textbf{0.42} && 2.4 \\
            
            \rownumber \rule{0pt}{3ex}
            & It started when I was 17.   
            & They'd make a cash drop,
            && 0.63 & 0.00 & \textbf{0.63} && 2.0  \\
            
            \rowcolor{gray!7}
            \rownumber \rule{0pt}{3ex}
            & A big nail should be put in your head    
            & Who are they
            && 0.74 & 0.00 & \textbf{0.74} && 1.2  \\

            \midrule

            \rownumber \rule{0pt}{3ex}
            & He told me so.
            & Oh, he did, huh?
            && 2.21 & 0.00 & \textbf{2.21} && 4.8 \\
            
            \rowcolor{gray!7}
            \rownumber \rule{0pt}{3ex}
            & There's a laundry. 
            & Have your clothes dry-cleaned, okay?
            && 0.81 & 2.89 & \textbf{3.70} && 4.4 \\
            
            \rownumber \rule{0pt}{3ex}
            & Then if I win, what are you going to do? 
            & When you win?
            && 1.04 & 7.01 & \textbf{8.05} && 4.2 \\
            
            \rowcolor{gray!7}
            \rownumber \rule{0pt}{3ex}
            & But what do you want me to do?
            & We want you to kick her off the team.
            && 10.20  & 1.53 & \textbf{11.72} && 5.0\\
            
            \bottomrule
        \end{tabular}
    \caption{Samples of utterance pairs scored with our method and human judgements (English). The scores of $\Sframe$ and $\Scontent$ were normalized by $\alpha$, $\beta$.}
    \label{tab:scored samples}
\end{table*}

% ==================================
\subsection{Results and Analysis}

Table~\ref{tab:scored correlation} shows the correlation between human scores and those automatically computed by each method.
Among the methods, $\Sours$ achieved the highest correlation with human scores.
Additionally, we also evaluated $\Sframe$ and $\Scontent$ as an ablation study of $\Sours$.
We found that both scores were less correlated than $\Sours$.
This result supports the hypothesis that both aspects, namely, connectivity and content relatedness, should be considered when evaluating the quality of utterance pairs.

Figure~\ref{fig:en_correlation-human-3} shows the distribution of automatically computed scores corresponding to human scores.%
    \footnote{See Appendix~\ref{a:en-correlation_dist} for the distributions of other methods.}
As shown in (c), $\Sours$ rarely overestimates utterance pairs with low human scores but underestimates those with high human scores.
The baseline methods presented in (a) and (b)  do not show such behavior. 
This behavior unique to $\Sours$ is safe for the noisy data filtering task since it can successfully detect lower-quality pairs with high precision.
On the other hand, improperly underestimating some acceptable pairs (i.e., low recall) is one downside of $\Sours$, and we discuss its influences in Section~\ref{ssec:LowRecall}.
We emphasize that $\Sours$ has a desirable property for noisy data filtering in today's situation where a sufficiently large corpus is available; it allows us to obtain a sufficient amount of clean data even if discarding a certain portion of potentially clean data. 
Interesting future work is to investigate how to improve our methods not to underestimate acceptable pairs while maintaining high precision.
It is nearly equivalent to develop an unsupervised approach of dialogue evaluation methods, and thus, this direction is a challenging and essential attempt.

Table~\ref{tab:scored samples} shows several examples of utterance pairs well-scored by $\Sframe$, $\Scontent$, and $\Sours$.
Note that the score ranges differ; e.g., human scores are in $[1, 5]$, while $\Scontent$ is in the range $[0, 1]$.%
    \footnote{See Appendix~\ref{a:score distributions} for score distributions on training data.}
Thus, we discuss relative score values; the comparison of absolute score values across the different methods would be meaningless.
These examples demonstrate that the complementary contributions of both $\Sframe$ and $\Scontent$ allow $\Sours$ to provide quality estimations close to human judgments.

% ==================================
\subsection{Discussion on Low Recall Property}
\label{ssec:LowRecall}

\paragraph{What types of pairs cause low recall?}
Since the proposed method prefers precision over recall, it tends to discard a certain number of acceptable utterance pairs during filtering.
To investigate the characteristics of such discarded (yet acceptable) pairs, we analyzed $27$ pairs.%
    \footnote{Some examples are listed in Appendix~\ref{a:post-hoc-analyses}}
These pairs were selected from those that obtained a human score of 4.0 or above ($77$ pairs) \emph{and} were among the worst $50$\% as scored by $\Sours$ ($100$ pairs).
Consequently, we found two potential issues.
One is that human annotators may sometimes easily find the connectivity or the content relatedness for the utterance pairs with the low $\Sours$ scores.
This observation indicates that $\Sframe$ and $\Scontent$ are still not perfect for scoring functions, and there remains room for improvement.
The possible drawbacks we have already noticed in $\Sframe$ and $\Scontent$ are that $\Sframe$ sometimes fails to capture the connectivity because of the limited coverage by a discrete phrase table-based approach, and $\Scontent$ is not robust for out-of-vocabulary of word vector.
The other case is that the human annotators gave high scores, but we found no connectivity and content relatedness in the utterance pairs.
We found that some utterance pairs without any connectivity and content relatedness can be judged as acceptable by the human annotators since they can imagine the underlying context and situation of the utterance pairs using human world knowledge, such as commonsense.
We think this is a challenging issue that exceeds our focus in this paper, and thus, remains as future work.

\begin{table}[t]
    \centering
    \small
    \begin{tabular}{l lccc}
         \toprule
         Scored data && len & distinct-1 & distinct-2 \\
         \midrule
         Top 50\% (remained)   && 9.02 & 0.028 & 0.472 \\
         Worst 50\% (removed)  && 9.00 & 0.030 & 0.470 \\
         \bottomrule
    \end{tabular}
    \caption{Comparison of the top and the worst utterance pairs' responses in the training data scored by our method (English).}
    \label{tab:scored-top-vs-worst-small}
\end{table}

\begin{table*}[!t]
    \centering
    \small
    \tabcolsep 7pt
    \setcount
    \renewcommand{\arraystretch}{1}
    \begin{tabular}{+l^c   ^r ^c^c HHH ^cHH HHH HHH   c cc}
        \toprule
        \multirow{2}{*}{Training data} &  \multirow{2}{*}{\# of pairs } & \multicolumn{15}{c}{Automatic evaluation} &  \multicolumn{3}{c}{Human evaluation} \\ \cmidrule{3-16} \cmidrule{18-20}
        & & len & distinct-1 & distinct-2 & BLEU-1$^{s}$ & bp$^{s}$ & B1bp$^{s}$ & BLEU-1 & bp & B1bp & METEOR & ROUGE & CID & EA & VE & GM  & Avg. & \xmark$^\downarrow$ & \cmark$^\uparrow$ \\
        \toprule
        non-filtered & 79,445,453 & 
        8.44 & 127/0.030 & 238/0.064 & 16.5 & 0.93 & 15.4 & 8.8 & 0.96 & 8.4 & 4.83 & 7.71 & 11.03 & 0.667 & 0.463 & 0.686 &
        3.37 & 38\,\% & 62\,\% \\

        \midrule

        \citet{csaky2019acl:filtering} SRC & 40,000,000 & 
        7.97 & 165/0.041 & 329/0.094 & 16.7 & 0.88 & 14.6 & 9.1 & 0.90 & 8.2 & 4.99 & 7.76 & 11.36 & 0.673 & 0.463 & 0.688 &
        3.56 & 25\,\% & 75\,\% \\
        \citet{csaky2019acl:filtering} TRG & 40,000,000 & 
        18.25 & 213/0.023 & 591/0.069 & 10.1 & 1.00 & 10.1 & 5.4 & 1.00 & 5.4 & 5.15 & 6.86 & 3.33 & 0.701 & 0.453 & 0.682 &
        2.85 & 65\,\% & 35\,\% \\
        \citet{junczys-dowmunt2018wmt:filtering} & 40,000,000 & 
        \textbf{8.63} & 206/0.048 & 478/0.125 & 17.0 & 0.95 & 16.2 & \textbf{9.4} & 0.98 & 9.2 & \textbf{5.16} & \textbf{8.32} & 10.25 & 0.668 & 0.463 & 0.688 &
        3.43 & 32\,\% & 68\,\% \\
        Ours $\Sours$ & 40,000,000 & 
        7.13 & \textbf{345/0.097} & \textbf{853/0.278} & 18.3 & 0.76 & 14.0 & \textbf{9.4} & 0.75 & 7.1 & 4.21 & 7.50 & 10.69 & 0.655 & 0.452 & 0.682 &
        3.73 & \textbf{15\,\%} & \textbf{85\,\%} \\
        
        \midrule
        Ours $\Sframe$ \scriptsize{(ablation study)} & 40,000,000 & 
        7.31 & 201/0.055 & 466/0.148 & 15.9 & 0.79 & 12.5 & 9.2 & 0.80 & 7.3 & 4.38 & 7.56 & 13.54 & 0.674 & 0.463 & 0.685 &
        3.69 & 19\,\% & 81\,\% \\
        Ours $\Scontent$ \scriptsize{(ablation study)} & 40,000,000 & 
        7.91 & 270/0.068 & 662/0.192 & 17.5 & 0.87 & 15.2 & \textbf{9.4} & 0.86 & 8.1 & 4.59 & 7.65 & 10.07 & 0.667 & 0.458 & 0.685 &
        \textbf{3.76} & 20\,\% & 80\,\% \\
        \midrule
        reference & &           
        9.04 & 1301/0.288 & 3244/0.807  
        & - & - & - & - & - & - & - & - & - & - & - & - 
        & - & - & - \\      
        \bottomrule
    \end{tabular}
    \caption{Evaluation results for generated responses (English; filtered out 50\%). \textbf{Bold} denotes the best results. The \xmark /\cmark \ shows the percentages of the low/high scored responses (i.e., human scores in $[1,3)$ or in $[3,5)$).}
    \label{tab:evaluation-results-en}
\end{table*}

\paragraph{Does our filtering undermine diversity?}
One might think that our method succeeds in filtering by assigning high scores to generic responses such as dull responses.
This concern makes sense since it is known that dialogue systems learned from the training data, including many generic utterances, tend to generate bland responses~\cite{csaky2019acl:filtering}.
To answer this interesting question, we confirmed the diversity of utterance pairs with a high score (i.e., remained as training data) and a low score (i.e., removed from training data) in our $\Sours$ (Table~\ref{tab:scored-top-vs-worst-small}).%
    \footnote{See Appendix~\ref{a:post-hoc-analyses} for more extensive result.}
As a result, there was no significant difference between them.
Therefore, we conclude that the proposed method does not prefer only generic responses and maintains the diversity of data.
It is an essential future attempt to improve the quality of dialogue data further (e.g., more diversity) after using the proposed method to remove unacceptable pairs.

\newcommand{\ra}{\scriptsize{A : } } % non-filtered
\newcommand{\rb}{\scriptsize{B : } } % ent-src
\newcommand{\rc}{\scriptsize{C : } } % ent-trg
\newcommand{\rd}{\scriptsize{D : } } % junczys-dowmunt2018wmt
\newcommand{\ro}{\scriptsize{Ours : } } % ours

\begin{table}[t!]
    \centering
    \small
    \setcount
    \renewcommand{\arraystretch}{1.15}
    \tabcolsep 2.5pt
    \begin{tabular}{ll}
        \toprule

        \rownumber 
        & \textbf{Well, I thought we'd wait for Mr. Oliver.} \\
        \multicolumn{2}{l}{
            \tabcolsep 2pt
            \begin{tabular}{rp{20.5em}}
                \ra & I don't think so. \\
                \rb & I'll be right back. \\
                \rc & I'm sorry, but I have to get back to work. \\
                \rd & No, I don't think so. \\
                \ro & Oh, I'm sorry. \\
            \end{tabular}
        } \\
        \midrule

        \rownumber
        & \textbf{Are you heading home?}\\
        \multicolumn{2}{l}{
            \tabcolsep 2pt
            \begin{tabular}{rp{20.5em}}
                \ra & No, I don't think so. \\
                \rb & No, I don't think so. \\
                \rc & No, no, no, no, no, no, no, no, no, no, no, no. \\
                \rd & No, I don't think so. \\
                \ro & No, I'm not. \\
            \end{tabular}
        } \\
        \midrule

        \rownumber
        & \textbf{Because we want to keep him safe.} \\
        \multicolumn{2}{l}{
            \tabcolsep 2pt
            \begin{tabular}{rp{20.5em}}
                \ra & I don't know what you're talking about. \\
                \rb & I don't know what you're talking about. \\
                \rc & I don't care if he's dead or alive or dead. \\
                \rd & I don't know what you're talking about. \\
                \ro & We need to get him out of here. \\
            \end{tabular}
        } \\
        \midrule

        \rownumber
        & \textbf{Quite soon, now, you'll be able to travel.} \\
        \multicolumn{2}{l}{
            \tabcolsep 2pt
            \begin{tabular}{rp{20.5em}}
                \ra & I don't think so. \\
                \rb & I'll be back in a minute. \\
                \rc & I'll be back in a few days, then I'll be back. \\
                \rd & I don't want to see you again. \\
                \ro & No, I'm not going anywhere. \\
            \end{tabular}
        } \\
        \midrule

        \rownumber
        & \textbf{Which is where exactly?} \\
        \multicolumn{2}{l}{
            \tabcolsep 2pt
            \begin{tabular}{rp{20.5em}}
                \ra & I don't know. \\
                \rb & I don't know. \\
                \rc & I don't know what you're talking about, but I don't know where you are. \\
                \rd & I don't know. \\
                \ro & I don't know. \\
            \end{tabular}
        } \\
        \bottomrule
    \end{tabular}
    \caption{Samples of input utterances \textbf{(bold)} and responses generated by models trained on filtered data by proposed method $\Sours$ (Ours). Other responses are generated by models trained on (A) non-filtered data and filtered data by baselines; (B) \citet{csaky2019acl:filtering}'s SRC method, (C) their TRG method, and (D) \citet{junczys-dowmunt2018wmt:filtering}'s method, respectively.}
    \label{tab:sample_response}
\end{table}

%%%%%%%%%%%%%%%%%%%%%%%%%%%%%%%%%%%%
\section{Experiments: Response Generation}
\label{sec:case study}
%%%%%%%%%%%%%%%%%%%%%%%%%%%%%%%%%%%%

This section reports on the effectiveness of the proposed method for filtering noisy data in neural response generation.

% ==================================
\subsection{Experimental Setup}

\paragraph{Training.}
\label{ssec:TrainingSettings}

We obtained the filtered training data $\mathcal{D}'$ by removing utterance pairs with low scores from the original dataset $\mathcal D$ (approximately $10$\% or $50$\% of total utterance pairs were removed).
As a response generation model, we used a Transformer~\cite{vaswani2017nips:transformer} based encoder-decoder model implemented in the \texttt{fairseq} toolkit~\cite{ott2019naacldemo:fairseq}.%
    \footnote{See Appendix~\ref{a:training settings} for training details.}
Transformer has demonstrated high performance in response generation~\cite{dinan2019iclr:wizard} and other NLP tasks.

\paragraph{Automatic evaluation.}
Here, we report the following metrics: the average response length in tokens (len), type-token ratio for $\{1,2\}$-grams (distinct-$\{1,2\}$), and and BLEU-1~\citep{Papineni2002acl:bleu}.
The latter was used as a reference-based metric; while it is widely used in previous studies~\cite{zhao2017acl:onetomany,baheti2018emnlp:generating-interesting-response,csaky2019acl:filtering}, some studies (e.g., ~\cite{liu2016emnlp:hownot}) have reported that BLEU-1 may not be highly correlated with the human evaluation of response generation.%
    \footnote{See Appendix~\ref{a:generated response with metrics} for more extensive evaluation results.}

\paragraph{Human evaluation.}
We evaluated the quality of the generated responses manually. 
We asked human evaluators recruited via Amazon Mechanical Turk to evaluate responses that are generated for $100$%
    \footnote{Same size as \citet{shen2017acl:conditional-variational,bao2020acl:plato}.}
input utterances randomly sampled from the test data. 
We used the same task setting and protocol as described in Section~\ref{ssec:scoring eval} to obtain the human scores for each pair. 
Higher human scores indicate that the better results.

% ==================================
\subsection{Results and Analysis}

Table~\ref{tab:evaluation-results-en} shows the results of automatic and human evaluations of the generated responses.
The model trained on the data filtered using the proposed method $\Sours$ produced more than three times as many distinct $\{1,2\}$-grams as the model trained on non-filtered data.
Furthermore, it outperformed the model trained on non-filtered data in the human evaluation, achieving the highest percentage of acceptable responses of 85\%. 
Additionally, these results of our $\Sours$ were better than other baselines.
To conclude, these experimental results indicate that the proposed scoring method can help generate diverse responses that are judged as acceptable by humans. 

This experiment provides empirical evidence for supporting our hypothesis that the performance of neural response generation models can be improved by just removing unacceptable utterance pairs from training data, which answers the research question formulated at the start of this paper.

%%%%%%%%%%%%%%%%%%%%%%%%%%%%%%%%%%%%
\section{Multilingual Availability}
\label{sec:Japanese}
%%%%%%%%%%%%%%%%%%%%%%%%%%%%%%%%%%%%

While the proposed method $\Sours$ was tested on an English corpus, it can potentially work for other languages as well. 
To demonstrate this, we selected Japanese dialogue data as another case study.%
    \footnote{See Appendix~\ref{a:japanese} for all experimental results on Japanese.} 
The linguistic phenomena in Japanese are quite different from those in English, thus making this experiment to be a good test of the applicability of the proposed method to non-English languages.

\paragraph{Japanese dataset.}
We prepare the Japanese dialogue data from Japanese OpenSubtitles~\cite{lison2018lrec:opensubtitles} containing roughly $3$M lines.
We obtain $1,\!893,\!477$ utterance pairs as our training data, which excludes our test and validation data.%
    \footnote{See Appendix~\ref{a:corpus_creation} for details on our data such as the preparation procedure and statistics.}

\begin{table}[!t]
    \centering
    \small
    \begin{tabular}{lrc}
        \toprule
        Scoring method & Spearman's $\rho$ & $p$-value \\
        \midrule
        \citet{csaky2019acl:filtering} SRC
        & $-$0.0553	
        & $4.4 \times 10^{-1}$\\
        \citet{csaky2019acl:filtering} TRG
        & $-$0.0366 
        & $6.1 \times 10^{-1}$ \\
        \citet{junczys-dowmunt2018wmt:filtering}
        & 0.1074 
        & $1.3 \times 10^{-1}$ \\
        Ours $\Sours$
        & \textbf{0.2491} 
        & $\mathbf{3.8 \times 10^{-4}}$ \\

        \midrule

        Ours $\Sframe$ \scriptsize{(ablation study)}
        & 0.1395	
        & $4.9 \times 10^{-2}$ \\
        Ours $\Scontent$ \scriptsize{(ablation study)}
        & 0.1504 
        & $3.3 \times 10^{-2}$ \\
        \bottomrule
    \end{tabular}
    \caption{Correlation coefficient between human scores and automatically computed scores (Japanese).}
    \label{tab:scored correlation_ja}
\end{table}

\begin{figure}[!t]
    \small
    \centering
    \includegraphics[trim = 0 10mm 0 0, width=0.7\columnwidth]{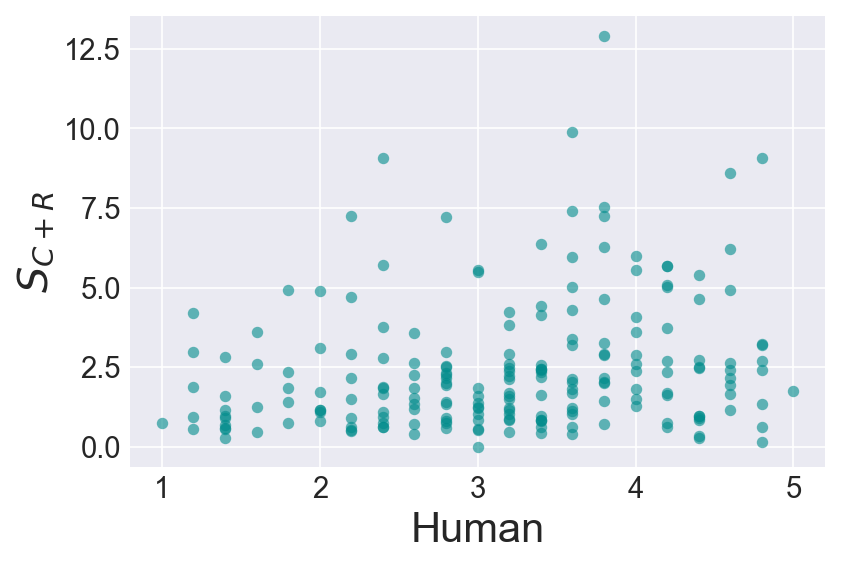}
    \caption{Distribution between human scores and our $\Sours$ (Japanese).}
    \label{fig:ja_correlation-human-onlyours}
\end{figure}

\begin{table*}[!t]
    \centering
    \small
    \tabcolsep 7pt
    \setcount
    \renewcommand{\arraystretch}{1}
    \begin{tabular}{+l^c   ^r ^c^c HHH ^cHH HHH HHH   c cc}
        \toprule
        \multirow{2}{*}{Training data} &  \multirow{2}{*}{\# of pairs } & \multicolumn{15}{c}{Automatic evaluation} &  \multicolumn{3}{c}{Human evaluation} \\ \cmidrule{3-16} \cmidrule{18-20}
        & & len & distinct-1 & distinct-2 & BLEU-1$^{s}$ & bp$^{s}$ & B1bp$^{s}$ & BLEU-1 & bp & B1bp & METEOR & ROUGE & CID & EA & VE & GM  & Avg. & \xmark$^\downarrow$ & \cmark$^\uparrow$ \\
        \toprule
        non-filterd & 1,893,477 & 
        5.91 & 268/0.091 & 509/0.207 & 14.0 & 0.79 & 11.1 & 13.4 & 0.79 & 10.6 & 5.44 & 10.98 & 16.27 & 0.723 & 0.438 & 0.585 &
        3.35 & 39\,\% & 61\,\% \\

        \midrule
        \citet{csaky2019acl:filtering} SRC & 1,700,000 & 
        5.75 & 295/0.102 & 550/0.231 & 13.9 & 0.77 & 10.6 & 13.2 & 0.76 & 10.0 & 5.09 & 10.79 & 15.03 & 0.711 & 0.430 & 0.575 &
        3.47 & 37\,\% & 63\,\%\\
        \citet{csaky2019acl:filtering} TRG & 1,700,000 & 
        \textbf{7.06} & 336/0.095 & 662/0.219 & 12.0 & 0.97 & 11.6 & 11.6 & 0.96 & 11.1 & \textbf{5.75} & 9.91 & 11.23 & 0.730 & 0.434 & 0.581 &
        3.37 & 34\,\% & 66\,\%\\
        \citet{junczys-dowmunt2018wmt:filtering} & 1,700,000 & 
        5.31 & 284/0.107 & 516/0.240 & 13.3 & 0.69 & 9.2 & 12.6 & 0.68 & 8.5 & 4.87 & 9.84 & 14.89 & 0.711 & 0.441 & 0.574 &
        3.46 & 32\,\% & 68\,\% \\
        Ours $\Sours$ & 1,700,000 & 
        5.68 & \textbf{319/0.112} & \textbf{582/0.249} & 14.4 & 0.75 & 10.8 & \textbf{13.9} & 0.75 & 10.5 & 5.42 & \textbf{11.22} & 17.22 & 0.725 & 0.441 & 0.585 &
        \textbf{3.61} & \textbf{27\,\%} & \textbf{73\,\%} \\
        
        \midrule
        Ours $\Sframe$ \scriptsize{(ablation study)}  & 1,700,000 & 
        5.51 & 264/0.096 & 492/0.218 & 14.4 & 0.72 & 10.4 & 13.7 & 0.72 & 9.8 & 5.28 & 10.74 & 15.43 & 0.724 & 0.447 & 0.586 &
        3.44 & 32\,\% & 68\,\% \\
        Ours $\Scontent$ \scriptsize{(ablation study)} & 1,700,000 & 
        5.73 & 296/0.103 & 555/0.234 & 13.2 & 0.76 & 10.1 & 12.5 & 0.76 & 9.5 & 5.20 & 9.85 & 12.82 & 0.719 & 0.441 & 0.579 &
        3.56 & 30\,\% & 70\,\% \\

        \midrule
        reference & & 
        7.29 & 750/0.206 & 1446/0.460  
        & - & - & - & - & - & - & - & - & - & - & - & -
        & - & - & - \\      
        \bottomrule
    \end{tabular}
    % \vskip -1mm
    \caption{Evaluation results for generated responses (Japanese; filtered out 10\%). \textbf{Bold} denotes the best results. The \xmark /\cmark \ shows the percentages of the low/high scored responses (i.e., human scores in $[1,3)$ or  in $[3,5)$).}
    \label{tab:evaluation-results-ja}
\end{table*}

% ==================================
\subsection{Data Scoring}
\label{ssec:JapaneseScoring}

\paragraph{Settings.}
To compute $\Sframe$, we defined a low co-occurrence frequency as less than 20, considering the size of the Japanese corpus, and consequently obtained the key phrase pairs $|\mathcal{P}|=19,\!992$.
To compute $\Scontent$, we used pre-trained fastText \cite{grave2018lrec:wordvec157} and learned common components from all sentences in the training data.

For human evaluation, we used Yahoo!~crowdsourcing%
    \footnote{\url{https://crowdsourcing.yahoo.co.jp/}}%
to hire native Japanese-speaking workers.
The task setting and protocol are the same as those for English (Section \ref{ssec:scoring eval}), regardless of the crowdsourcing platform.

\paragraph{Results and analysis.}
Table~\ref{tab:scored correlation_ja} shows the correlation between human scores and those automatically computed by each method.
Our method $\Sours$ has the highest correlation with human scores, although the overall result is lower than that obtained for the English dataset.
Figure~\ref{fig:ja_correlation-human-onlyours} shows the distribution of our $\Sours$ corresponding to human scores.
Similar to the result obtained for English as presented in Figure~\ref{fig:en_correlation-human-3} (c), $\Sours$ rarely overestimates utterance pairs with low human scores but underestimates those with high human scores in Japanese.

% ==================================
\subsection{Response Generation}
\label{ssec:JapaneseFiltering}

\paragraph{Settings.}
We used the same experimental settings described in Section~\ref{ssec:TrainingSettings} for the preparation of filtered data $\mathcal{D}'$ and model training.

\paragraph{Results and analysis.}
Table~\ref{tab:evaluation-results-ja} shows the results of evaluations of the generated responses.
The filtered data generated by $\Sours$ provided the best results in terms of almost all the metrics, including human evaluation.
It supports our hypothesis that the proposed method is also suitable for non-English languages.

%%%%%%%%%%%%%%%%%%%%%%%%%%%%%%%%%%%%
\section{Relationship with Evaluation Metric}
\label{sec:relatedWork}
%%%%%%%%%%%%%%%%%%%%%%%%%%%%%%%%%%%%

The proposed method $\Sours$ maps an utterance pair to a score (scalar value) in terms of the quality of dialogue.
That is, formally, our method is similar to the reference-free automatic evaluation metrics for dialogue agents; both of them evaluate the response given an input utterance and also map into a score. 
Recently, the novel reference-free metrics for evaluating generated responses such as USR~\cite{Mehri2020acl:usr} or \textsc{MaUde}~\cite{sinha2020acl:maude} ware developed.
While it is possible to use them as a scoring method for filtering noisy data, in theory, there are some concerns with applying them in practice.
One is the difference of the data of interest; since evaluation metrics are intended for responses generated as dialogue, i.e., somewhat valid dialogue data, it is unclear whether they also work for apparently noisy data.
Another one is the difference of desired properties; evaluation metrics need to be sensitive to ``how good is it?''\ while the filtering requires to detect ``is it a dialogue?''\ with high accuracy.
It would be interesting to investigate the effectiveness of reference-free metrics for noisy dialogue data filtering tasks, and vice versa.
We leave these investigations for future work.

In contrast, reference-based metrics require a reference response (i.e., ground truth) when they calculate scores; such metrics include the traditional overlap-based BLEU, ROUGE, METEOR, embedding-based metrics~\cite{liu2016emnlp:hownot}, and neural network-based RUBER~\cite{tao2018aaai:ruber} and ADEM~\cite{lowe2017acl:adem}
Thus, these methods cannot straightforwardly be considered as alternatives to the proposed method, which aims at filtering.

%%%%%%%%%%%%%%%%%%%%%%%%%%%%%%%%%%%%
\section{Conclusion}
\label{sec:conclusion}
%%%%%%%%%%%%%%%%%%%%%%%%%%%%%%%%%%%%

In light of the success of noisy corpus filtering in neural machine translation, we attempted to filter out unacceptable utterance pairs from large dialogue corpora in an unsupervised manner.
The proposed scoring method estimates the quality of utterance pairs by focusing on the two crucial aspects of dialogue, namely, the \emph{connectivity} and \emph{content relatedness} of utterance pairs.
We demonstrated that our scoring method has a higher correlation with human judgment than recently proposed methods.
Furthermore, we provided empirical evidence that our method improves the performance of a response generation model by removing unacceptable utterance pairs from its training data. 
We hope that this study will facilitate discussions in the dialogue response generation community regarding the issue of filtering noisy corpora.

\clearpage
%%%%%%%%%%%%%%%%%%%%%%%%%%%%%%%%%%%%
\section*{Acknowledgments}
%%%%%%%%%%%%%%%%%%%%%%%%%%%%%%%%%%%%
This work was supported by JSPS KAKENHI Grant Numbers JP19H04162, JP19J21913.

%%%%%%%%%%%%%%%%%%%%%%%%%%%%%%%%%%%%
%%%%%%%%%%%%%%%%%%%%%%%%%%%%%%%%%%%%
% reference
%%%%%%%%%%%%%%%%%%%%%%%%%%%%%%%%%%%%
%%%%%%%%%%%%%%%%%%%%%%%%%%%%%%%%%%%%
\bibliography{Mendeley}

\begin{thebibliography}{53}
\expandafter\ifx\csname natexlab\endcsname\relax\def\natexlab#1{#1}\fi

\bibitem[{Adiwardana et~al.(2020)Adiwardana, Luong, So, Hall, Fiedel,
  Thoppilan, Yang, Kulshreshtha, Nemade, Lu~Quoc, and
  Le}]{adiwardana2020arxiv:meena}
Daniel Adiwardana, Minh-Thang Luong, David~R So, Jamie Hall, Noah Fiedel, Romal
  Thoppilan, Zi~Yang, Apoorv Kulshreshtha, Gaurav Nemade, Yifeng Lu~Quoc, and
  V~Le. 2020.
\newblock \href {https://arxiv.org/pdf/2001.09977.pdf} {{Towards a Human-like
  Open-Domain Chatbot}}.
\newblock In \emph{aiXiv preprint arXiv:2001.09977}.

\bibitem[{Akama et~al.(2017)Akama, Inada, Inoue, Kobayashi, and
  Inui}]{akama2017ijcnlp:generating}
Reina Akama, Kazuaki Inada, Naoya Inoue, Sosuke Kobayashi, and Kentaro Inui.
  2017.
\newblock \href {https://www.aclweb.org/anthology/I17-2069} {{Generating
  Stylistically Consistent Dialog Responses with Transfer Learning}}.
\newblock In \emph{Proceedings of the 8th International Joint Conference on
  Natural Language Processing (IJCNLP)}, volume~2, pages 408--412.

\bibitem[{Arora et~al.(2017)Arora, Liang, and Ma}]{arora2017iclr:sif}
Sanjeev Arora, Yingyu Liang, and Tengyu Ma. 2017.
\newblock \href {https://openreview.net/forum?id=SyK00v5xx} {{A Simple but
  Tough-to-Beat Baseline for Sentence Embeddings}}.
\newblock In \emph{5th International Conference on Learning Representations
  (ICLR)}.

\bibitem[{Baheti et~al.(2018)Baheti, Ritter, Li, and
  Dolan}]{baheti2018emnlp:generating-interesting-response}
Ashutosh Baheti, Alan Ritter, Jiwei Li, and Bill Dolan. 2018.
\newblock \href {https://doi.org/10.18653/v1/D18-1431} {{Generating More
  Interesting Responses in Neural Conversation Models with Distributional
  Constraints}}.
\newblock In \emph{Proceedings of the 2018 Conference on Empirical Methods in
  Natural Language Processing (EMNLP)}, pages 3970--3980.

\bibitem[{Bao et~al.(2020)Bao, He, Wang, Wu, and Wang}]{bao2020acl:plato}
Siqi Bao, Huang He, Fan Wang, Hua Wu, and Haifeng Wang. 2020.
\newblock \href {https://doi.org/10.18653/v1/2020.acl-main.9} {{PLATO:
  Pre-trained Dialogue Generation Model with Discrete Latent Variable}}.
\newblock In \emph{Proceedings of the 58th Annual Meeting of the Association
  for Computational Linguistics (ACL)}, pages 85--96.

\bibitem[{Bojanowski et~al.(2017)Bojanowski, Grave, Joulin, and
  Mikolov}]{Bojanowski2017tacl:fasttext}
Piotr Bojanowski, Edouard Grave, Armand Joulin, and Tomas Mikolov. 2017.
\newblock \href {https://doi.org/10.1162/tacl{\_}a{\_}00051} {{Enriching Word
  Vectors with Subword Information}}.
\newblock \emph{Transactions of the Association for Computational Linguistics
  (TACL)}, 5:135--146.

\bibitem[{Bouma(2009)}]{bouma2009gscl:npmi}
Gerlof Bouma. 2009.
\newblock \href
  {https://svn.spraakdata.gu.se/repos/gerlof/pub/www/Docs/npmi-pfd.pdf}
  {{Normalized (Pointwise) Mutual Information in Collocation Extraction}}.
\newblock In \emph{Proceedings of the International Conference of the German
  Society for Computational Linguistics and Language Technology (GSCL)}, pages
  31--40.

\bibitem[{Cho and May(2020)}]{cho2020acl:yesand}
Hyundong Cho and Jonathan May. 2020.
\newblock \href {https://doi.org/10.18653/v1/2020.acl-main.218} {{Grounding
  Conversations with Improvised Dialogues}}.
\newblock In \emph{Proceedings of the 58th Annual Meeting of the Association
  for Computational Linguistics (ACL)}, pages 2398--2413.

\bibitem[{Conneau et~al.(2017)Conneau, Kiela, Schwenk, Barrault, and
  Bordes}]{Conneau2017emnlp:universalrepresentation}
Alexis Conneau, Douwe Kiela, Holger Schwenk, Loic Barrault, and Antoine Bordes.
  2017.
\newblock \href {https://doi.org/10.1.1.156.2685} {{Supervised Learning of
  Universal Sentence Representations from Natural Language Inference Data}}.
\newblock In \emph{Proceedings of the 2017 Conference on Empirical Methods in
  Natural Language Processing (EMNLP)}, pages 670--680.

\bibitem[{Cs{\'{a}}ky et~al.(2019)Cs{\'{a}}ky, Purgai, and
  Recski}]{csaky2019acl:filtering}
Richárd Cs{\'{a}}ky, Patrik Purgai, and Gábor Recski. 2019.
\newblock \href {https://doi.org/10.18653/v1/P19-1567} {{Improving Neural
  Conversational Models with Entropy-Based Data Filtering}}.
\newblock In \emph{Proceedings of the 57th Annual Meeting of the Association
  for Computational Linguistics (ACL)}, pages 5650--5669.

\bibitem[{Dinan et~al.(2019)Dinan, Roller, Shuster, Fan, Auli, and
  Weston}]{dinan2019iclr:wizard}
Emily Dinan, Stephen Roller, Kurt Shuster, Angela Fan, Michael Auli, and Jason
  Weston. 2019.
\newblock \href {https://openreview.net/forum?id=r1l73iRqKm} {{Wizard of
  Wikipedia: Knowledge-Powered Conversational Agents}}.
\newblock In \emph{7th International Conference on Learning Representations
  (ICLR)}.

\bibitem[{Dyer et~al.(2013)Dyer, Chahuneau, and
  Smith}]{dyer2013naacl:fastalign}
Chris Dyer, Victor Chahuneau, and Noah~A Smith. 2013.
\newblock \href {https://www.aclweb.org/anthology/N13-1073} {{A Simple, Fast,
  and Effective Reparameterization of IBM Model 2}}.
\newblock In \emph{Proceedings of the 2013 Conference of the North American
  Chapter of the Association for Computational Linguistics: Human Language
  Technologies (NAACL-HLT)}, pages 644--648.

\bibitem[{Galley et~al.(2015)Galley, Brockett, Sordoni, Ji, Auli, Quirk,
  Mitchell, Gao, and Dolan}]{galley2015aclijcnlp:deltableu}
Michel Galley, Chris Brockett, Alessandro Sordoni, Yangfeng Ji, Michael Auli,
  Chris Quirk, Margaret Mitchell, Jianfeng Gao, and Bill Dolan. 2015.
\newblock \href {https://doi.org/10.3115/v1/P15-2073} {{deltaBLEU: A
  Discriminative Metric for Generation Tasks with Intrinsically Diverse
  Targets}}.
\newblock In \emph{Proceedings of the 53rd Annual Meeting of the Association
  for Computational Linguistics and the 7th International Joint Conference on
  Natural Language Processing (ACL-IJCNLP)}, volume~2, pages 445--450.

\bibitem[{Grave et~al.(2018)Grave, Bojanowski, Gupta, Joulin, and
  Mikolov}]{grave2018lrec:wordvec157}
Edouard Grave, Piotr Bojanowski, Prakhar Gupta, Armand Joulin, and Tomas
  Mikolov. 2018.
\newblock \href {https://www.aclweb.org/anthology/L18-1550} {{Learning Word
  Vectors for 157 Languages}}.
\newblock In \emph{Proceedings of the 11th International Conference on Language
  Resources and Evaluation (LREC)}, pages 3483--3487.

\bibitem[{Henderson et~al.(2019)Henderson, Budzianowski, Casanueva, Coope,
  Gerz, Kumar, Mrk{\v{s}}i´c, Spithourakis, Su, Vuli´c, and
  Wen}]{henderson2019convai:convdata}
Matthew Henderson, Paweł Budzianowski, Iñigo Casanueva, Sam Coope, Daniela
  Gerz, Girish Kumar, Nikola~Mrkši´c Mrk{\v{s}}i´c, Georgios Spithourakis,
  Pei-Hao Su, Ivan~Vuli´c Vuli´c, and Tsung-Hsien Wen. 2019.
\newblock \href {https://www.aclweb.org/anthology/W19-4101} {{A Repository of
  Conversational Datasets}}.
\newblock In \emph{Proceedings of the First Workshop on NLP for Conversational
  AI}, pages 1--10.

\bibitem[{Junczys-Dowmunt(2018)}]{junczys-dowmunt2018wmt:filtering}
Marcin Junczys-Dowmunt. 2018.
\newblock \href {https://doi.org/10.18653/v1/W18-6478} {{Dual Conditional
  Cross-Entropy Filtering of Noisy Parallel Corpora}}.
\newblock In \emph{Proceedings of the 3rd Conference on Machine Translation:
  Shared Task Papers (WMT)}, pages 888--895.

\bibitem[{Koehn et~al.(2007)Koehn, Hoang, Birch, Callison-Burch, Federico,
  Bertoldi, Cowan, Shen, Mit, Zens, Aachen, Dyer, Bojar, and
  Cornell}]{koehen2017acl:moses}
Philipp Koehn, Hieu Hoang, Alexandra Birch, Chris Callison-Burch, Marcello
  Federico, Nicola Bertoldi, Brooke Cowan, Wade Shen, Christine~Moran Mit,
  Richard Zens, Rwth Aachen, Chris Dyer, Ondřej Bojar, and Evan~Herbst
  Cornell. 2007.
\newblock \href {https://www.aclweb.org/anthology/P07-2045.pdf} {{Moses: Open
  Source Toolkit for Statistical Machine Translation}}.
\newblock In \emph{Proceedings of the 45th Annual Meeting of the Association
  for Computational Linguistics Companion (ACL) Volume Proceedings of the Demo
  and Poster Sessions}, pages 177--180.

\bibitem[{Koehn et~al.(2018)Koehn, Khayrallah, Heafield, and
  Forcada}]{koehn2018wmt:filteringfindings}
Philipp Koehn, Huda Khayrallah, Kenneth Heafield, and Mikel~L Forcada. 2018.
\newblock \href {https://doi.org/10.18653/v1/W18-6453} {{Findings of the WMT
  2018 Shared Task on Parallel Corpus Filtering}}.
\newblock In \emph{Proceedings of the 3rd Conference on Machine Translation:
  Shared Task Papers (WMT)}, pages 726--739.

\bibitem[{Koehn and Knowles(2017)}]{koehn2017nmt:sixchallenges}
Philipp Koehn and Rebecca Knowles. 2017.
\newblock \href {https://www.aclweb.org/anthology/W17-3204.pdf} {{Six
  Challenges for Neural Machine Translation}}.
\newblock In \emph{Proceedings of the First Workshop on Neural Machine
  Translation}, pages 28--39.

\bibitem[{Li et~al.(2016)Li, Galley, Brockett, Gao, and
  Dolan}]{li2016naacl:diversity}
Jiwei Li, Michel Galley, Chris Brockett, Jianfeng Gao, and Bill Dolan. 2016.
\newblock \href {https://doi.org/10.18653/v1/N16-1014} {{A Diversity-Promoting
  Objective Function for Neural Conversation Models}}.
\newblock In \emph{Proceedings of the 2016 Conference of the North American
  Chapter of the Association for Computational Linguistics: Human Language
  Technologies (NAACL-HLT)}, pages 110--119.

\bibitem[{Likert(1932)}]{likert1932archivesofpsycho:scale}
Rensis Likert. 1932.
\newblock \href {https://psycnet.apa.org/record/1933-01885-001} {{A technique
  for the measurement of attitudes}}.
\newblock \emph{Archives of psychology}, 22(140):1--55.

\bibitem[{Lison and Tiedemann(2016)}]{lison2016lrec:opensubtitles}
Pierre Lison and Jörg Tiedemann. 2016.
\newblock \href {https://www.aclweb.org/anthology/L16-1147/}
  {{OpenSubtitles2016: Extracting Large Parallel Corpora from Movie and TV
  Subtitles}}.
\newblock In \emph{Proceedings of the 10th International Conference on Language
  Resources and Evaluation (LREC)}, pages 923--929.

\bibitem[{Lison et~al.(2018)Lison, Tiedemann, and
  Kouylekov}]{lison2018lrec:opensubtitles}
Pierre Lison, Jörg Tiedemann, and Milen Kouylekov. 2018.
\newblock \href {https://www.aclweb.org/anthology/L18-1275}
  {{OpenSubtitles2018: Statistical Rescoring of Sentence Alignments in Large,
  Noisy Parallel Corpora}}.
\newblock In \emph{Proceedings of the 11th International Conference on Language
  Resources and Evaluation (LREC)}, pages 1742--1748.

\bibitem[{Liu et~al.(2016)Liu, Lowe, Serban, Noseworthy, Charlin, and
  Pineau}]{liu2016emnlp:hownot}
Chia-Wei Liu, Ryan Lowe, Iulian~V Serban, Michael Noseworthy, Laurent Charlin,
  and Joelle Pineau. 2016.
\newblock \href {https://www.aclweb.org/anthology/D16-1230.pdf} {{How NOT To
  Evaluate Your Dialogue System: An Empirical Study of Unsupervised Evaluation
  Metrics for Dialogue Response Generation}}.
\newblock In \emph{Proceedings of the 2016 Conference on Empirical Methods in
  Natural Language Processing (EMNLP)}, pages 2122--2132.

\bibitem[{Lowe et~al.(2017{\natexlab{a}})Lowe, Noseworthy, Serban, A-Gontier,
  Bengio, and Pineau}]{lowe2017acl:adem}
Ryan Lowe, Michael Noseworthy, Iulian~V Serban, Nicolas A-Gontier, Yoshua
  Bengio, and Joelle Pineau. 2017{\natexlab{a}}.
\newblock \href {https://doi.org/10.18653/v1/P17-1103} {{Towards an Automatic
  Turing Test: Learning to Evaluate Dialogue Responses}}.
\newblock In \emph{Proceedings of the 55th Annual Meeting of the Association
  for Computational Linguistics (ACL)}, pages 1116--1126.

\bibitem[{Lowe et~al.(2017{\natexlab{b}})Lowe, Noseworthy, Serban,
  Angelard-Gontier, Bengio, and Pineau}]{lowe2017acl:autoturingtest}
Ryan Lowe, Michael Noseworthy, Iulian~Vlad Serban, Nicolas Angelard-Gontier,
  Yoshua Bengio, and Joelle Pineau. 2017{\natexlab{b}}.
\newblock \href {https://doi.org/10.18653/v1/P17-1103} {{Towards an Automatic
  Turing Test: Learning to Evaluate Dialogue Responses}}.
\newblock In \emph{Proceedings of the 55th Annual Meeting of the Association
  for Computational Linguistics (ACL)}, volume~1, pages 1116--1126.

\bibitem[{Marelli et~al.(2014{\natexlab{a}})Marelli, Bentivogli, Baroni,
  Bernardi, Menini, and
  Zamparelli}]{marelli2014semeval:compositional-distributional-model}
Marco Marelli, Luisa Bentivogli, Marco Baroni, Raffaella Bernardi, Stefano
  Menini, and Roberto Zamparelli. 2014{\natexlab{a}}.
\newblock \href {https://doi.org/10.3115/v1/S14-2001} {{SemEval-2014 Task 1:
  Evaluation of Compositional Distributional Semantic Models on Full Sentences
  through Semantic Relatedness and Textual Entailment}}.
\newblock In \emph{Proceedings of the 8th International Workshop on Semantic
  Evaluation (SemEval)}, pages 1--8.

\bibitem[{Marelli et~al.(2014{\natexlab{b}})Marelli, Menini, Baroni,
  Bentivogli, Bernardi, and Zamparelli}]{marelli2014lrec:sick}
Marco Marelli, Stefano Menini, Marco Baroni, Luisa Bentivogli, Raffaella
  Bernardi, and Roberto Zamparelli. 2014{\natexlab{b}}.
\newblock \href {https://www.aclweb.org/anthology/L14-1314/} {{A SICK cure for
  the evaluation of compositional distributional semantic models}}.
\newblock In \emph{Proceedings of the 9th International Conference on Language
  Resources and Evaluation (LREC)}, pages 216--223.

\bibitem[{Mehri and Eskenazi(2020)}]{Mehri2020acl:usr}
Shikib Mehri and Maxine Eskenazi. 2020.
\newblock \href {https://www.aclweb.org/anthology/2020.acl-main.64/} {{USR: An
  Unsupervised and Reference Free Evaluation Metric for Dialog Generation}}.
\newblock In \emph{Proceedings of the 58th Annual Meeting of the Association
  for Computational Linguistics (ACL)}, pages 681--707.

\bibitem[{Mikolov et~al.(2018)Mikolov, Grave, Bojanowski, Puhrsch, and
  Joulin}]{mikolov2018lrec:advances-in-wordrep}
Tomas Mikolov, Edouard Grave, Piotr Bojanowski, Christian Puhrsch, and Armand
  Joulin. 2018.
\newblock \href {https://www.aclweb.org/anthology/L18-1008} {{Advances in
  Pre-Training Distributed Word Representations}}.
\newblock In \emph{Proceedings of the 11th International Conference on Language
  Resources and Evaluation (LREC)}, pages 52--55.

\bibitem[{Morishita et~al.(2018)Morishita, Suzuki, and
  Nagata}]{morishita2018wmt:ntt}
Makoto Morishita, Jun Suzuki, and Masaaki Nagata. 2018.
\newblock \href {https://doi.org/10.18653/v1/W18-6421} {{NTT’s Neural Machine
  Translation Systems for WMT 2018}}.
\newblock In \emph{Proceedings of the 3rd Conference on Machine Translation:
  Shared Task Papers (WMT)}, pages 461--466.

\bibitem[{Ott et~al.(2019)Ott, Edunov, Baevski, Fan, Gross, Ng, Grangier, and
  Auli}]{ott2019naacldemo:fairseq}
Myle Ott, Sergey Edunov, Alexei Baevski, Angela Fan, Sam Gross, Nathan Ng,
  David Grangier, and Michael Auli. 2019.
\newblock \href {https://doi.org/10.18653/v1/N19-4009} {{fairseq: A Fast,
  Extensible Toolkit for Sequence Modeling}}.
\newblock In \emph{Proceedings of the 2019 Conference of the North American
  Chapter of the Association for Computational Linguistics (Demonstrations)
  (NAACL-HLT)}, pages 48--53.

\bibitem[{Papineni et~al.(2002)Papineni, Roukos, Ward, and
  Zhu}]{Papineni2002acl:bleu}
Kishore Papineni, Salim Roukos, Todd Ward, and Wei-Jing Zhu. 2002.
\newblock \href {https://www.aclweb.org/anthology/P02-1040.pdf} {{BLEU: a
  Method for Automatic Evaluation of Machine Translation}}.
\newblock In \emph{Proceedings of the 40th Annual Meeting of the Association
  for Computational Linguistics (ACL)}, pages 311--318.

\bibitem[{Pei and Li(2018)}]{pei2018emnlp:s2spmn}
Jiaxin Pei and Chenliang Li. 2018.
\newblock \href {https://doi.org/10.18653/v1/D18-1082} {{S2SPMN: A Simple and
  Effective Framework for Response Generation with Relevant Information}}.
\newblock In \emph{Proceedings of the 2018 Conference on Empirical Methods in
  Natural Language Processing (EMNLP)}, pages 745--750.

\bibitem[{Ritter et~al.(2011)Ritter, Cherry, and Dolan}]{ritter2011:datadriven}
Alan Ritter, Colin Cherry, and William~B Dolan. 2011.
\newblock \href {https://www.aclweb.org/anthology/D11-1054} {{Data-Driven
  Response Generation in Social Media}}.
\newblock In \emph{Proceedings of the 2011 Conference on Empirical Methods in
  Natural Language Processing (EMNLP)}, pages 583--593.

\bibitem[{Sacks(1989)}]{sacks1989humanstudies:conversationalrule}
Harvey Sacks. 1989.
\newblock \href {http://www.jstor.org/stable/20009058} {{Lecture One: Rules of
  Conversational Sequence}}.
\newblock \emph{Human Studies}, 12(3/4):217--233.

\bibitem[{Sedoc et~al.(2019)Sedoc, Ippolito, Kirubarajan, Thirani, Ungar, and
  Callison-Burch}]{sedoc2019naacldemo:chateval}
João Sedoc, Daphne Ippolito, Arun Kirubarajan, Jai Thirani, Lyle Ungar, and
  Chris Callison-Burch. 2019.
\newblock \href {https://doi.org/10.18653/v1/N19-4011} {{ChatEval: A Tool for
  Chatbot Evaluation}}.
\newblock In \emph{Proceedings of the 2019 Conference of the North American
  Chapter of the Association for Computational Linguistics (NAACL)
  (Demonstrations)}, pages 60--65.

\bibitem[{Sennrich et~al.(2016)Sennrich, Haddow, and
  Birch}]{sennrich2016acl:subword}
Rico Sennrich, Barry Haddow, and Alexandra Birch. 2016.
\newblock \href {https://doi.org/10.18653/v1/P16-1162} {{Neural Machine
  Translation of Rare Words with Subword Units}}.
\newblock In \emph{Proceedings of the 54th Annual Meeting of the Association
  for Computational Linguistics (ACL)}, volume~1, pages 1715--1725.

\bibitem[{Sennrich and Zhang(2019)}]{sennrich2019acl:nmtcasestudy}
Rico Sennrich and Biao Zhang. 2019.
\newblock \href {https://www.aclweb.org/anthology/P19-1021/} {{Revisiting
  Low-Resource Neural Machine Translation: A Case Study}}.
\newblock In \emph{Proceedings of the 57th Annual Meeting of the Association
  for Computational Linguistics (ACL)}, pages 211--221.

\bibitem[{Shang et~al.(2015)Shang, Lu, and
  Li}]{shang2015aclijcnlp:neuralresponding}
Lifeng Shang, Zhengdong Lu, and Hang Li. 2015.
\newblock \href {https://doi.org/10.3115/v1/P15-1152} {{Neural Responding
  Machine for Short-Text Conversation}}.
\newblock In \emph{Proceedings of the 53rd Annual Meeting of the Association
  for Computational Linguistics and the 7th International Joint Conference on
  Natural Language Processing (ACL-IJCNLP)}, volume~1, pages 1577--1586.

\bibitem[{Shang et~al.(2018)Shang, Fu, Peng, Feng, Zhao, and
  Yan}]{shang2018ijcai:calibration}
Mingyue Shang, Zhenxin Fu, Nanyun Peng, Yansong Feng, Dongyan Zhao, and Rui
  Yan. 2018.
\newblock \href {https://www.ijcai.org/Proceedings/2018/603} {{Learning to
  Converse with Noisy Data: Generation with Calibration}}.
\newblock In \emph{Proceedings of the Twenty-Seventh International Joint
  Conference on Artificial Intelligence (IJCAI-18)}, pages 4338--4344.

\bibitem[{Shen et~al.(2017)Shen, Su, Li, Li, Niu, Zhao, Aizawa, and
  Long}]{shen2017acl:conditional-variational}
Xiaoyu Shen, Hui Su, Yanran Li, Wenjie Li, Shuzi Niu, Yang Zhao, Akiko Aizawa,
  and Guoping Long. 2017.
\newblock \href {https://doi.org/10.18653/v1/P17-2080} {{A Conditional
  Variational Framework for Dialog Generation}}.
\newblock In \emph{Proceedings of the 55th Annual Meeting of the Association
  for Computational Linguistics (ACL)}, volume~2, pages 504--509.

\bibitem[{Sidnell(2010)}]{sidnell2010book:conversation}
Jack Sidnell. 2010.
\newblock \href {https://books.google.co.jp/books?id=-Wfx_uM-bmQC}
  {\emph{{Conversation Analysis: An Introduction}}}.
\newblock Language in Society. John Wiley {\&} Sons.

\bibitem[{Sinha et~al.(2020)Sinha, Parthasarathi, Wang, Lowe, Hamilton, and
  Pineau}]{sinha2020acl:maude}
Koustuv Sinha, Prasanna Parthasarathi, Jasmine Wang, Ryan Lowe, William~L
  Hamilton, and Joelle Pineau. 2020.
\newblock \href {https://www.aclweb.org/anthology/2020.acl-main.220/}
  {{Learning an Unreferenced Metric for Online Dialogue Evaluation}}.
\newblock In \emph{Proceedings of the 58th Annual Meeting of the Association
  for Computational Linguistics}, pages 2430--2441. Association for
  Computational Linguistics.

\bibitem[{Subramanian et~al.(2018)Subramanian, Trischler, Bengio, and
  Pal}]{subramanian2018iclr:generalpurpose}
Sandeep Subramanian, Adam Trischler, Yoshua Bengio, and Christopher~J Pal.
  2018.
\newblock \href {https://openreview.net/forum?id=B18WgG-CZ} {{Learning General
  Purpose Distributed Sentence Representations via Large Scale Multi-task
  Learning}}.
\newblock In \emph{6th International Conference on Learning Representations
  (ICLR)}.

\bibitem[{Sutskever et~al.(2014)Sutskever, Vinyals, and
  Le}]{Sutskever2014nips:seq2seq}
Ilya Sutskever, Oriol Vinyals, and Quoc~V. Le. 2014.
\newblock \href
  {http://papers.nips.cc/paper/5346-sequence-to-sequence-learning-with-neural-networks}
  {{Sequence to Sequence Learning with Neural Networks}}.
\newblock In \emph{Advances in Neural Information Processing Systems 27
  (NIPS)}, pages 3104--3112.

\bibitem[{Tao et~al.(2018)Tao, Mou, Zhao, and Yan}]{tao2018aaai:ruber}
Chongyang Tao, Lili Mou, Dongyan Zhao, and Rui Yan. 2018.
\newblock \href {https://arxiv.org/abs/1701.03079} {{RUBER: An Unsupervised
  Method for Automatic Evaluation of Open-Domain Dialog Systems}}.
\newblock In \emph{Proceedings of the 32nd AAAI Conference on Artificial
  Intelligence (AAAI-18)}.

\bibitem[{Vaswani et~al.(2017)Vaswani, Shazeer, Parmar, Uszkoreit, Jones,
  Gomez, Kaiser, and Polosukhin}]{vaswani2017nips:transformer}
Ashish Vaswani, Noam Shazeer, Niki Parmar, Jakob Uszkoreit, Llion Jones,
  Aidan~N. Gomez, Łukasz Kaiser, and Illia Polosukhin. 2017.
\newblock \href {https://papers.nips.cc/paper/7181-attention-is-all-you-need}
  {{Attention is All you Need}}.
\newblock In \emph{Advances in Neural Information Processing Systems 30
  (NIPS)}, pages 5998--6008.

\bibitem[{Vinyals and Le(2015)}]{vinyals2015icml:neuralconv}
Oriol Vinyals and Quoc Le. 2015.
\newblock \href {http://arxiv.org/pdf/1506.05869v3.pdf} {{A Neural
  Conversational Model}}.
\newblock In \emph{Proceedings of the 31st International Conference on Machine
  Learning (ICML) Deep Learning Workshop}.

\bibitem[{Xing et~al.(2017)Xing, Wu, Wu, Liu, Huang, Zhou, and
  Ma}]{xing2017aaai:topicaware}
Chen Xing, Wei Wu, Yu~Wu, Jie Liu, Yalou Huang, Ming Zhou, and Wei-Ying Ma.
  2017.
\newblock \href {http://aaai.org/ocs/index.php/AAAI/AAAI17/paper/view/14563}
  {{Topic Aware Neural Response Generation}}.
\newblock In \emph{Proceedings of the 31st AAAI Conference on Artificial
  Intelligence}, pages 3351--3357.

\bibitem[{Xu et~al.(2018{\natexlab{a}})Xu, Du{\v{s}}ek, Konstas, and
  Rieser}]{xu2018emnlp:filtering}
Xinnuo Xu, Ondře Du{\v{s}}ek, Ioannis Konstas, and Verena Rieser.
  2018{\natexlab{a}}.
\newblock \href {https://doi.org/10.18653/v1/D18-1432} {{Better Conversations
  by Modeling, Filtering, and Optimizing for Coherence and Diversity}}.
\newblock In \emph{Proceedings of the 2018 Conference on Empirical Methods in
  Natural Language Processing (EMNLP)}, pages 3981--3991.

\bibitem[{Xu et~al.(2018{\natexlab{b}})Xu, Jiang, Liu, Rong, Wu, Wang, Wang,
  and Wang}]{xu2018naacl:lsdscc}
Zhen Xu, Nan Jiang, Bingquan Liu, Wenge Rong, Bowen Wu, Baoxun Wang, Zhuoran
  Wang, and Xiaolong Wang. 2018{\natexlab{b}}.
\newblock \href {https://doi.org/10.18653/v1/N18-1188} {{LSDSCC: a Large Scale
  Domain-Specific Conversational Corpus for Response Generation with Diversity
  Oriented Evaluation Metrics}}.
\newblock In \emph{Proceedings of the 2018 Conference of the North American
  Chapter of the Association for Computational Linguistics: Human Language
  Technologies (NAACL-HLT)}, volume~1, pages 2070--2080.

\bibitem[{Zhao and Eskenazi(2017)}]{zhao2017acl:onetomany}
Tiancheng Zhao and Maxine Eskenazi. 2017.
\newblock \href {https://doi.org/10.18653/v1/P17-1061} {{Learning
  Discourse-level Diversity for Neural Dialog Models using Conditional
  Variational Autoencoders}}.
\newblock In \emph{Proceedings of the 55th Annual Meeting of the Association
  for Computational Linguistics}, pages 654--664.

\end{thebibliography}
\bibliographystyle{acl_natbib}

%%%%%%%%%%%%%%%%%%%%%%%%%%%%%%%%%%%%
%%%%%%%%%%%%%%%%%%%%%%%%%%%%%%%%%%%%
%  Appendix 
%%%%%%%%%%%%%%%%%%%%%%%%%%%%%%%%%%%%
%%%%%%%%%%%%%%%%%%%%%%%%%%%%%%%%%%%%
\appendix
\clearpage

\setlist{itemsep=0pt, topsep=1pt}
\setlist[itemize]{leftmargin=*}
\setlist[enumerate]{leftmargin=*}

%%%%%%%%%%%%%%%%%%%%%%%%%%%%%%%%%%%%
\section{Preliminary Experiment Settings} 
\label{a:preliminary experiment}
%%%%%%%%%%%%%%%%%%%%%%%%%%%%%%%%%%%%

\paragraph{Dataset.}
For our preliminary experiment (Section~\ref{sec:introduction}), we use OpenSubtitles~\cite{lison2018lrec:opensubtitles} in English, one of the largest corpora of movie scripts that are freely available and has been used in many data-driven dialogue response generations.
We automatically obtained dialogue paired-data from the corpus which does not contain speaker annotations on the dialogue turns \textit{following the previous methods}~\cite{vinyals2015icml:neuralconv,li2016naacl:diversity}.
Specifically, we extracted the consecutive two lines as an utterance pair based on the assumption that each line corresponds to a full-speaker's turn.
We collected pairs from the dataset in which the length of the utterance and response was $3$-$25$ words each and obtained the dialogue dataset.
For counting the number of words, we used SpaCy\footnote{\url{https://spacy.io/}} to tokenize each utterance and response.

\paragraph{Evaluation settings.}
We used Amazon Mechanical Turk (MTurk) to evaluate the data manually.
In our experiments, randomly sampled $100$ utterance pairs were evaluated by native English speakers.
We filtered out unreliable workers by integrating attention checks.
We requested five workers to evaluate each pair with a five-point Likert scale ($5$: Strongly agree to $1$: Strongly disagree)~\cite{likert1932archivesofpsycho:scale} as an answer to the following question: \textit{Is the sequence of the two utterances acceptable as a dialogue?}.

\paragraph{Result.}
As a result of our preliminary experiment, we discover that, out of all scores given for utterance pairs, 25\% was unacceptable (scored as $1$: Strongly disagree or $2$: Disagree) and almost half was acceptable (scored as $5$: Strongly agree or $4$: Agree).
The inter-annotator agreement (Krippendorff's alpha) was $0.33$.

%%%%%%%%%%%%%%%%%%%%%%%%%%%%%%%%%%%%
\section{Dialogue Data Constructions}
\label{a:corpus_creation}
%%%%%%%%%%%%%%%%%%%%%%%%%%%%%%%%%%%%

\paragraph{English Dataset.}

In our experiments (Section~\ref{sec:experiments}, \ref{sec:case study}), we used the OpenSubtitles%
    \footnote{\url{http://opus.nlpl.eu/OpenSubtitles-v2018.php}}%
~\citep{lison2018lrec:opensubtitles} as an example of the noisy million-scales English dialogue corpus.
In addition to the previous method for the extraction of pair data (as described in Appedix~\ref{a:preliminary experiment}), we cleaned the data with some heuristic preprocesses.
Some processings were inspired by the technique of noisy-parallel corpus filtering on NMT fields.
The additional preprocesses that we conducted are as follows:
\begin{itemize}
    \item Using languid%
        \footnote{\url{https://github.com/saffsd/langid.py}}%
        , which is a tool that detects the language for given sentences, we removed the utterance pairs judged as any language other than the target language.
    \item Removed the parrot-back utterance pairs.
    \item Removed duplicate utterance pairs in order to remove the completely repeated conversational sequences, such as the opening scenes of serial dramas.
\end{itemize}
Eventually, we obtained $79,\!621,\!506$ utterance pairs as our English dialogue corpus.
For our experiments, we divided them into training/validation/test data.
Table~\ref{tab:corpus size en} shows the statistics of our English dataset.
The ``\# pairs'' indicates the number of utterance pairs obtained by the previous method described in Appedix~\ref{a:preliminary experiment}.

\begin{table}[!h]
    \centering
    \small
    \tabcolsep 3.5pt
    \begin{tabular}{r rrrr}
        \toprule
        \textbf{Data} & \# works & \# lines   & \# pairs      & \# our pairs  \\
        \midrule
        Corpus & 446,612   & 441,452,475   & 230,597,913             & 79,621,506    \\
        \midrule
        Train  & 442,433   & 441,065,310   & 230,392,431   & 79,445,453    \\
        Valid  & 200	    & 195,297	    & 104,007       & 90,317        \\
        Test   & 200	    & 191,868	    & 101,475       & 85,736        \\
        \bottomrule
    \end{tabular}
    \caption{The statistics of our English dataset.}
    \label{tab:corpus size en}
\end{table}

\paragraph{Japanese Dataset.}

For our other experiment (Section~\ref{sec:Japanese}), we prepared our Japanse corpus from OpenSubtitles~\citep{lison2018lrec:opensubtitles}.
The data construction process, including preprocesses, is the same as those for English (as described in Appendix~\ref{a:preliminary experiment} and the previous paragraph).
We used mecab%
    \footnote{\url{https://taku910.github.io/mecab/}}
to tokenize the Japanese utterances.
Eventually, we obtained $1,\!917,\!721$ utterance pairs as our Japanese dialogue corpus.
For our experiments, we divided them into training/validation/test data.
Table~\ref{tab:corpus size ja} shows the statistics of our Japanese dataset.
The ``\# pairs'' indicates the number of utterance pairs obtained by the previous method described in Appedix~\ref{a:preliminary experiment}.

\begin{table}[!h]
    \centering
    \small
    \tabcolsep 5pt
    \begin{tabular}{r rrrr}
        \toprule
        \textbf{Data} & \# works & \# lines   & \# pairs      & \# our pairs  \\
        \midrule
        Corpus: & 3,546	    & 3,170,155     & 2,266,127             & 1,917,721             \\
        \midrule
        Train  & 3,506     & 3,135,812     & 2,240,847     & 1,893,477     \\
        Valid  & 20	    & 15,489	    & 11,939        & 11,486        \\
        Test   & 20        & 18,854	    & 13,341        & 12,758        \\
        \bottomrule
    \end{tabular}
    \caption{The statistics of our Japanese dataset.}
    \label{tab:corpus size ja}
\end{table}

%%%%%%%%%%%%%%%%%%%%%%%%%%%%%%%%%%%%
\section{Experimental Details of Proposed Method}
\label{a: proposed method}
%%%%%%%%%%%%%%%%%%%%%%%%%%%%%%%%%%%%

\subsection{Computing Connectivity as $\Sframe$}
To compute alignment points with fastAlign, we set the null alignment probability to $0.5$ and used the `grow-diag-final' heuristics.
To extract phrase pairs with Moses using the information of alignment points, we used the following settings: alignment=`grow-diag-final-and', reordering=`msd-bidirectional-fe', and first-step=$4$.
Furthermore, we extended the standard phrase extraction algorithm (Algorithm~\ref{alg: phrase pair extraction}) to only extract phrases that have at least one alignment point for every row and column when considering the matrix view of phrases (Algorithm~\ref{alg: modified phrase pair extraction}). 
This is because unaligned words should not be positively dealt with in the evaluation of connectivity.

\subsection{Computing Content relatedness as $\Scontent$}
For SIF weighting in Japanese data, we obtained word frequency data from jawiki dataset%
    \footnote{\url{https://dumps.wikimedia.org/jawiki/}}%
following English word frequency data provided in the author's implementation.%
    \footnote{\url{https://github.com/PrincetonML/SIF}}

%%%%%%%%%%%%%%%%%%%%%%%%%%%%%%%%%%%%
\section{Training Details for Response Generation Model}
\label{a:training settings}
%%%%%%%%%%%%%%%%%%%%%%%%%%%%%%%%%%%%

To obtain the response generation model, we used a Transformer~\cite{vaswani2017nips:transformer} based encoder-decoder model implemented in the \texttt{fairseq} toolkit%
    \footnote{\url{https://github.com/pytorch/fairseq}}%
~\cite{ott2019naacldemo:fairseq}.
We used `\texttt{--arch transformer\_wmt\_en\_de\_big}' option with its default configuration, and set the number of maximum training steps to $100$K. 
We used the byte pair encoding%
 \footnote{\url{https://github.com/rsennrich/subword-nmt}}%
~\cite{sennrich2016acl:subword} for token segmentation and set its vocabulary size to $16$K.
The numbers of parameters in our models were roughly $223$M.
We trained our models on $8$ NVIDIA DGX-$1$ Tesla V$100$ GPUs. 
It took approximately $6$ hours for training one model.

\begin{algorithm}[]
    \footnotesize
    \textbf{Input:} word alignment $\mathcal{A}$ for sentence pair $(x,y)$\\
    \textbf{Output:} set of phrase pair $\overline{\mathcal{P}}$ 
    \begin{algorithmic}[1]
        \For{$f_\mathrm{start} \xleftarrow{} 1, \cdots, \LEN{x}$}
            \For{$f_\mathrm{end} \xleftarrow{} f_\mathrm{start}, \cdots, \LEN{x}$}
                \State $e_\mathrm{start} \xleftarrow{} \LEN{y}$
                \State $e_\mathrm{end} \xleftarrow{} 0$
                \ForAll{$(f,e) \in \mathcal{A}$}
                    \If{$f_\mathrm{start} \leq f \leq f_\mathrm{end}$}
                        \State $e_\mathrm{start} \xleftarrow{} \mathrm{min}(e, e_\mathrm{start})$
                        \State $e_\mathrm{end} \xleftarrow{} \mathrm{max}(e, e_\mathrm{end})$
                    \EndIf
                \EndFor
                \State add \texttt{extract}($e_\mathrm{start}, e_\mathrm{end}, f_\mathrm{start},f_\mathrm{end}$) to set $\overline{\mathcal{P}}$
            \EndFor
        \EndFor
    \end{algorithmic}

    \textbf{function} \texttt{extract}($e_\mathrm{start}, e_\mathrm{end}, f_\mathrm{start},f_\mathrm{end}$)
    \begin{algorithmic}[1]
        \State \Return \{\} \textbf{if} $e_\mathrm{end} = 0$
        \ForAll{$(e,f) \in \mathcal{A}$}
            \State \Return \{\} \textbf{if} $f < f_\mathrm{start}$ or $f > f_\mathrm{end}$
        \EndFor
        \State $F = \{\}$
        \State $e_\mathrm{s} \xleftarrow{} e_\mathrm{start}$
        \Repeat 
            \State $e_\mathrm{e} \xleftarrow{} e_\mathrm{end}$
            \Repeat
                \State add phrase pair $(f_\mathrm{start} ... f_\mathrm{end}, e_\mathrm{s} ... e_\mathrm{e})$ to set $F$
                \State $e_\mathrm{e}++$
            \Until $e_\mathrm{e}$ aligned
            \State $e_\mathrm{s}--$
        \Until $e_\mathrm{s}$ aligned
        \State \Return $F$
    \end{algorithmic}
    \caption{Phrase pair extraction}
    \label{alg: phrase pair extraction}
\end{algorithm}

\begin{algorithm}[]
    \footnotesize
    \textbf{Input:} word alignment $\mathcal{A}$ for sentence pair $(x,y)$\\
    \textbf{Output:} set of phrase pair $\overline{\mathcal{P}}$ 
    \begin{algorithmic}[1]
        \For{$f_\mathrm{start} \xleftarrow{} 1, \cdots, \LEN{x}$}
            \For{$f_\mathrm{end} \xleftarrow{} f_\mathrm{start}, \cdots, \LEN{x}$}
                \State $e_\mathrm{start} \xleftarrow{} \LEN{y}$
                \State $e_\mathrm{end} \xleftarrow{} 0$
                \State $\mathcal{F} =\{\}$
                \State $\mathcal{E} =\{\}$
                \ForAll{$(f,e) \in \mathcal{A}$}
                    \If{$f_\mathrm{start} \leq f \leq f_\mathrm{end}$}
                        \State $e_\mathrm{start} \xleftarrow{} \mathrm{min}(e, e_\mathrm{start})$
                        \State $e_\mathrm{end} \xleftarrow{} \mathrm{max}(e, e_\mathrm{end})$
                        \State $\mathcal{F} \xleftarrow{} \mathcal{F} \cup \{f\}$
                        \State $\mathcal{E} \xleftarrow{} \mathcal{E} \cup \{e\}$
                        \EndIf
                \EndFor
                \State add \texttt{extract}($e_\mathrm{start}, e_\mathrm{end}, f_\mathrm{start},f_\mathrm{end}, \mathcal{F}, \mathcal{E}$) to set $\overline{\mathcal{P}}$
            \EndFor
        \EndFor
    \end{algorithmic}

    \textbf{function} \texttt{extract}($e_\mathrm{start}, e_\mathrm{end}, f_\mathrm{start},f_\mathrm{end}, \mathcal{F}, \mathcal{E}$)
    \begin{algorithmic}[1]
        \State \Return \{\} \textbf{if} $e_\mathrm{end} = 0$
        \For{$f \xleftarrow{} f_\mathrm{start}, \cdots, f_\mathrm{end}$}
            \State \Return \{\} \textbf{if} $f \notin \mathcal{F}$
        \EndFor
        \For{$e \xleftarrow{} e_\mathrm{start}, \cdots, e_\mathrm{end}$}
            \State \Return \{\} \textbf{if} $e \notin \mathcal{E}$
        \EndFor
        \State \Return  phrase pair $(f_\mathrm{start} ... f_\mathrm{end}, e_\mathrm{start} ... e_\mathrm{end})$
    \end{algorithmic}
    \caption{Modified phrase pair extraction}
    \label{alg: modified phrase pair extraction}
\end{algorithm}

%%%%%%%%%%%%%%%%%%%%%%%%%%%%%%%%%%%%
\section{Experimental Results on English}
%%%%%%%%%%%%%%%%%%%%%%%%%%%%%%%%%%%%

% ==================================
\subsection{Distributions between Human and Automatic Scoring}
\label{a:en-correlation_dist}

\begin{figure*}[]
    \centering
    \small
    \begin{tabular}{c}
        \begin{minipage}{0.33\hsize}
            \centering
            \includegraphics[trim = 0 3mm 0 0, width=0.97\columnwidth]{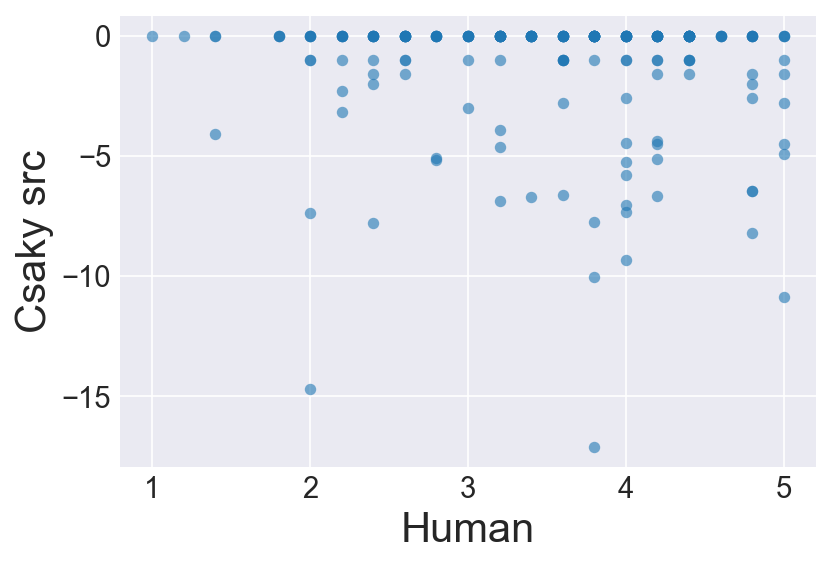}
            \hspace{20mm} (a) \quad \citet{csaky2019acl:filtering} SRC
        \end{minipage}
        \begin{minipage}{0.33\hsize}
            \centering
            \includegraphics[trim = 0 3mm 0 0, width=0.97\columnwidth]{en-correlation-human-pairscore-entropy-trg.png}
            \hspace{20mm} (b) \quad \citet{csaky2019acl:filtering} TRG
        \end{minipage}
        \begin{minipage}{0.33\hsize}
            \centering
            \includegraphics[trim = 0 3mm 0 0, width=0.97\columnwidth]{en-correlation-human-pairscore-dualconditional-crossent.png}
            \hspace{20mm} (c) \quad \citet{junczys-dowmunt2018wmt:filtering}
        \end{minipage}
        \\ \\
        \begin{minipage}{0.33\hsize}
            \centering
            \includegraphics[trim = 0 3mm 0 0, width=0.97\columnwidth]{en-correlation-human-pairscore-ours.png}
            \hspace{20mm}  (d) \quad Ours $\Sours$
        \end{minipage}
        \begin{minipage}{0.33\hsize}
            \centering
            \includegraphics[trim = 0 3mm 0 0, width=0.97\columnwidth]{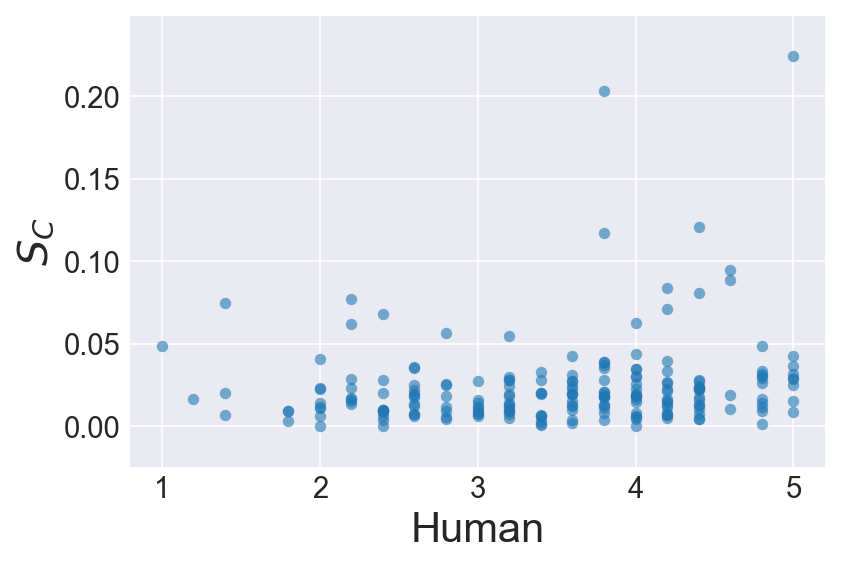}
            \hspace{20mm}  (e) \quad Ours $\Sframe$
        \end{minipage}
        \begin{minipage}{0.33\hsize}
            \centering
            \includegraphics[trim = 0 3mm 0 0, width=0.97\columnwidth]{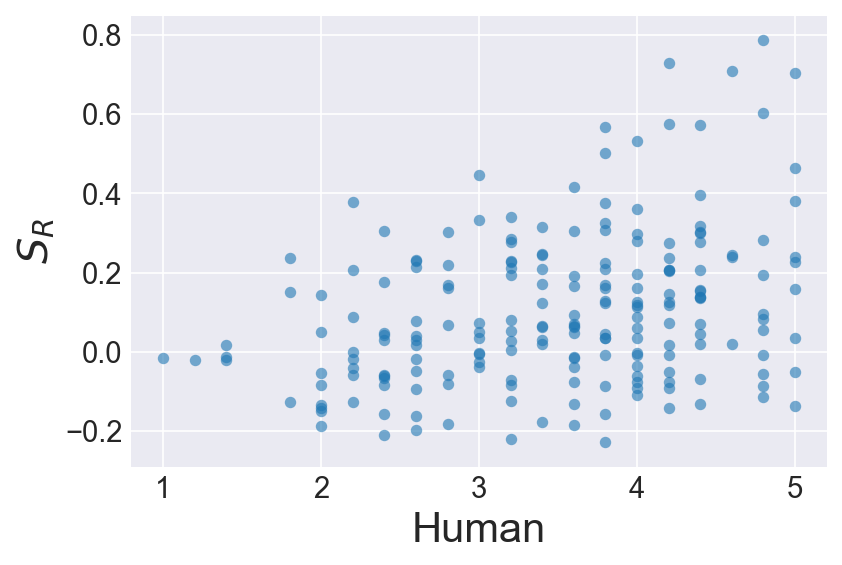}
            \hspace{20mm} (f) \quad Ours $\Scontent$
        \end{minipage}
    \end{tabular}
    \caption{Distributions between human scores and automatically computed scores by each method (English).}
    \label{fig:en_correlation-human-all}
\end{figure*}

Figure~\ref{fig:en_correlation-human-all} shows that, for all the models including ablations, the distributions between human scores and automatically computed scores.

\begin{figure*}[]
    \centering
    \begin{tabular}{c}
        \begin{minipage}{0.33\hsize}
            \centering
            \includegraphics[width=0.95\columnwidth]{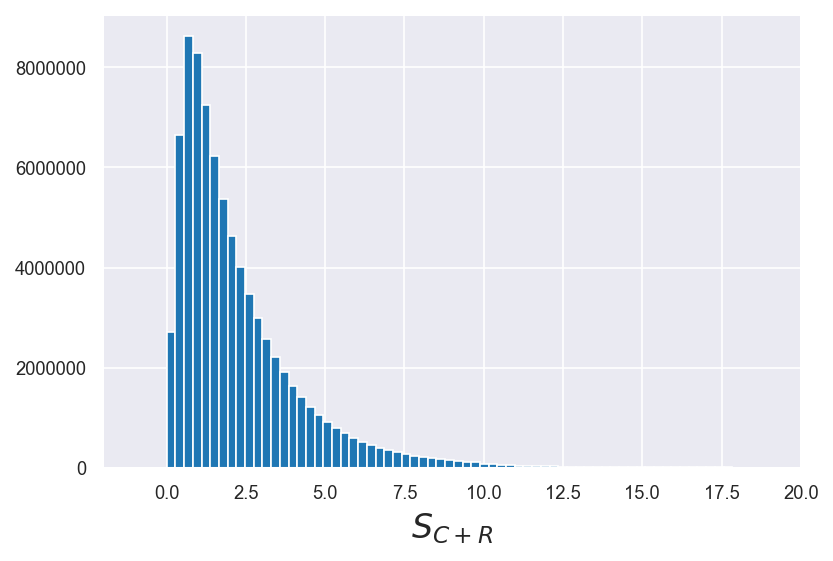}
            \hspace{1.6cm} (a) $\Sours (x,y)$
            \label{fig:s_ours en}
        \end{minipage}
        \begin{minipage}{0.33\hsize}
            \centering
            \includegraphics[width=0.95\columnwidth]{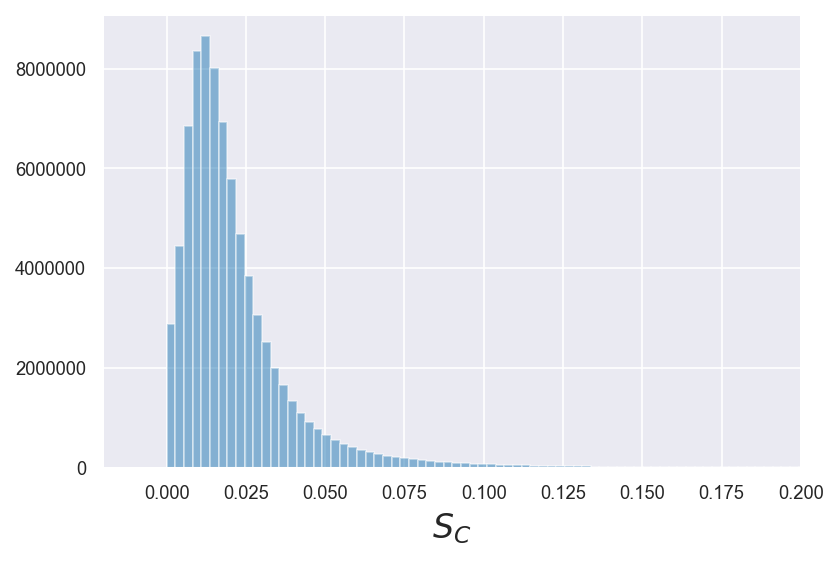}
            \hspace{1.6cm} (b) $\Sframe (x,y)$
            \label{fig:s_frame en}
        \end{minipage}
        \begin{minipage}{0.33\hsize}
            \centering
            \includegraphics[width=0.95\columnwidth]{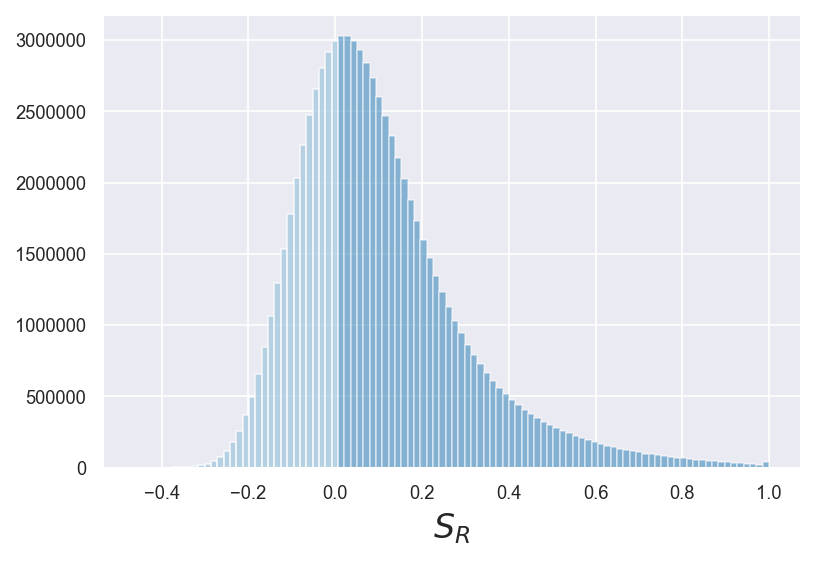}
            \hspace{1.6cm} (c) $\Scontent (x,y)$
            \label{fig:s_content en}
        \end{minipage}
    \end{tabular}
    \caption{Score distributions of our $\Sframe$, $\Scontent$, $\Sours$ across our training data (English).}
    \label{fig:scores en}
\end{figure*}

\begin{table*}[]
    \centering
    \small
    \tabcolsep 3.5pt
        \setcount
        \begin{tabular}{rp{38ex}p{35ex} l rrr l c}
            \toprule
            & Utterance & Response && $\Sframe$\, & $\Scontent$ & $\Sours$ && Human \\
            \cmidrule{1-3} \cmidrule{5-7} \cmidrule{9-9}
            
            % \rowcolor{gray!7}
            \rownumber \rule{0pt}{3ex}
            & What happened to your hand?
            & Just a scratch.
            && 1.38 & 0.00 & \textbf{1.38} && 4.8  \\
            % Sframe 力不足、取ってくれ〜〜〜 1
            
            \rowcolor{gray!7}
            \rownumber \rule{0pt}{3ex}
            & But Carcharodontosaurus has the more lethal bite.
            & This time, the Spinosaurus triumphed.
            && 0.22 & 0.68 & \textbf{1.72} && 4.0 \\
            % Scontent 力不足、未知語:Carcharodontosaurus,Spinosaurus 1
 
            % \rowcolor{gray!7}
            \rownumber \rule{0pt}{3ex}
            & I'm right here with you.	
            & Come on, boys.
            && 1.39 & 0.00 & \textbf{1.39} && 4.0 \\
            % 2観点ないがOK（dull? wildcard?） 2

            \rowcolor{gray!7}
            \rownumber \rule{0pt}{3ex}
            & What Is It Officer Chan?
            & Brother Ho, I must leave now,
            && 1.04 & 0.68 & \textbf{1.72} && 4.4 \\
            % 2観点なくてもOK（dull? wildcard?） 2

            % \rowcolor{gray!7}
            \rownumber \rule{0pt}{3ex}
            & Out on the balcony.
            & You shouldn't have come.
            && 0.30 & 0.00 & \textbf{0.30} && 4.2 \\
            % ここ以外の文脈が必要 2
            % これスパイダーマンだったりしない…？？？？？
            
            \bottomrule
        \end{tabular}
        % \vskip -1mm
    \caption{Samples of utterance pairs that cause low recall scored with our method and human judgements (English). The scores of $\Sframe$ and $\Scontent$ were normalized by $\alpha$, $\beta$.}
    \label{tab:scored samples low recall}
\end{table*}

\begin{table*}[]
    \centering
    \small
    \tabcolsep 4pt
    \begin{tabular}{r lccc lccc lccc}
         \toprule
         \multirow{2}{*}{Scored data} && \multicolumn{3}{c}{Utterance pair} && \multicolumn{3}{c}{Utterance (source-side)} && \multicolumn{3}{c}{Response (target-side)} \\
         \cmidrule{3-5} \cmidrule{7-9} \cmidrule{11-13}
         && len & distinct-1 & distinct-2 && len & distinct-1 & distinct-2 && len & distinct-1 & distinct-2 \\
         \midrule
         Top 50\% (remained)   && 18.06 & 0.018 & 0.313 && 9.05 & 0.028 & 0.474 && 9.02 & 0.028 & 0.472 \\
         Worst 50\% (removed) && 17.92 & 0.019 & 0.316 && 8.92 & 0.030 & 0.476 && 9.00 & 0.030 & 0.470 \\
         \midrule
         Top 90\% (remained)  && 17.99 & 0.013 & 0.229 && 9.00 & 0.022 & 0.382 && 8.99 & 0.022 & 0.378 \\
         Worst 10\% (removed)&& 17.99 & 0.047 & 0.654 && 8.85 & 0.068 & 0.872 && 9.14 & 0.067 & 0.873 \\
         \bottomrule
    \end{tabular}
    \caption{Comparison of the top and the worst utterance pairs in the training data scored by our method (English).}
    \label{tab:scored-top-vs-worst}
\end{table*}

\begin{table*}[]
    \centering
    \scriptsize
    \tabcolsep 5pt
    \begin{tabular}{+l^r|^r ^c^c HHH^c^c^c ^c^c^c ^c^c^c}
        \toprule
        English & \# of pairs 
        & len & dist1 & dist2 & B1 & bp & B1bp & B1 & bp & B1bp & MET & ROU & CID & EA & VE & GM \\
        \toprule
        non-filtered & 79,445,453 & 
        8.44 & 127/0.030 & 238/0.064 & 16.5 & 0.93 & 15.4 & 8.8 & 0.96 & 8.4 & 4.83 & 7.71 & 11.03 & 0.667 & 0.463 & 0.686 \\

        \midrule
        \scriptsize{\textbf{Filtered out 10\%:}}  \\
        \citet{csaky2019acl:filtering} SRC & 70,000,000 & 
        8.59 & 122/0.028 & 222/0.058 & 16.7 & 0.95 & 15.8 & 9.3 & 0.98 & 9.1 & 5.38 & 8.17 & 12.48 & 0.680 & 0.466 & 0.691\\
        \citet{csaky2019acl:filtering} TRG & 70,000,000 & 
        16.73 & 194/0.023 & 507/0.064 & 10.8 & 1.00 & 10.8 & 6.0 & 1.00 & 6.0 & 5.63 & 7.25 & 4.11 & 0.699 & 0.440 & 0.683\\
        \citet{junczys-dowmunt2018wmt:filtering} & 70,000,000 & 
        8.91 & 126/0.028 & 225/0.057 & 16.2 & 0.99 & 16.0 & 8.9 & 1.00 & 8.9 & 5.12 & 7.68 & 8.55 & 0.673 & 0.466 & 0.688\\
        Ours $\Sours$ & 70,000,000 & 
        8.43 & 183/0.043 & 403/0.108 & 16.4 & 0.93 & 15.3 & 9.2 & 0.95 & 8.8 & 4.95 & 7.92 & 10.26 & 0.674 & 0.462 & 0.687\\
        Ours $\Sframe$ & 70,000,000 & 
        8.60 & 130/0.030 & 231/0.061 & 16.3 & 0.95 & 15.5 & 9.1 & 0.99 & 9.0 & 5.11 & 7.95 & 10.53 & 0.682 & 0.467 & 0.688\\
        Ours $\Scontent$ & 70,000,000 & 
        8.42 & 155/0.037 & 306/0.083 & 16.8 & 0.93 & 15.6 & 9.2 & 0.95 & 8.7 & 4.93 & 7.89 & 8.76 & 0.664 & 0.464 & 0.687\\

        \midrule
        \scriptsize{\textbf{Filtered out 50\%:}}  \\
        \citet{csaky2019acl:filtering} SRC & 40,000,000 & 
        7.97 & 165/0.041 & 329/0.094 & 16.7 & 0.88 & 14.6 & 9.1 & 0.90 & 8.2 & 4.99 & 7.76 & 11.36 & 0.673 & 0.463 & 0.688\\
        \citet{csaky2019acl:filtering} TRG & 40,000,000 & 
        18.25 & 213/0.023 & 591/0.069 & 10.1 & 1.00 & 10.1 & 5.4 & 1.00 & 5.4 & 5.15 & 6.86 & 3.33 & 0.701 & 0.453 & 0.682\\
        \citet{junczys-dowmunt2018wmt:filtering} & 40,000,000 & 
        8.63 & 206/0.048 & 478/0.125 & 17.0 & 0.95 & 16.2 & 9.4 & 0.98 & 9.2 & 5.16 & 8.32 & 10.25 & 0.668 & 0.463 & 0.688\\
        Ours $\Sours$ & 40,000,000 & 
        7.13 & 345/0.097 & 853/0.278 & 18.3 & 0.76 & 14.0 & 9.4 & 0.75 & 7.1 & 4.21 & 7.50 & 10.69 & 0.655 & 0.452 & 0.682\\
        Ours $\Sframe$ & 40,000,000 & 
        7.31 & 201/0.055 & 466/0.148 & 15.9 & 0.79 & 12.5 & 9.2 & 0.80 & 7.3 & 4.38 & 7.56 & 13.54 & 0.674 & 0.463 & 0.685\\
        Ours $\Scontent$ & 40,000,000 & 
        7.91 & 270/0.068 & 662/0.192 & 17.5 & 0.87 & 15.2 & 9.4 & 0.86 & 8.1 & 4.59 & 7.65 & 10.07 & 0.667 & 0.458 & 0.685\\

        \midrule
        reference &            
        & 9.04 & 1301/0.288 & 3244/0.807 & - & - & - & - & - & - & - & - & - & - & - & - \\
        \bottomrule
    \end{tabular}
    \caption{Automatic evaluation results for generated responses (English).
    BLEU-1 (B1) and its brief penalty (bp), ROUGE (ROU)$\times 100$, METEOR$\times 100$ (MET), CIDEr$\times 100$ (CID). Embedding-based metrics: Embedding Average Cosine Similarity (EA), Vector Extrema Cosine Similarity (VE), Greedy Matching (GM).}
    \label{tab:autmatic_evaluations}
\end{table*}

% ====================================
\subsection{Score Distributions} 
\label{a:score distributions}

Figure~\ref{fig:scores en} shows that the score distributions of our $\Sframe$, $\Scontent$, $\Sours$ across our training data.
Note that, for computing $\Sours$, $\Scontent$ that less than $0$ (pale -colored part in (c)) are treated as $0$ (Equation~\ref{eq:s_content}).

% ====================================
\subsection{Qualitative Analysis for Low Recall} 
\label{a:post-hoc-analyses}

Table~\ref{tab:scored samples low recall} shows the samples of utterance pairs that cause low recall property.
In the $1$st and $2$nd pairs, humans can observe the connectivity or the content relatedness, but $\Sframe$ or $\Scontent$ failed to provide high scores.
For example, ``Carcharodontosaurus'' and ``Spinosaurus'' were unknown words for $\Scontent$.
Other pairs cannot be correctly determined from only these two perspectives.
The $3$rd and $4$th pairs are acceptable, although they have neither the connectivity nor the content relatedness. 
The $5$th pair needs more external contexts and knowledge to determine whether it is acceptable as dialogue.

% ====================================
\subsection{Comparison of Top versus Worst Data} 

Table~\ref{tab:scored-top-vs-worst} shows that the comparison of utterance pairs with a high score (i.e., remained as training data; top $50$\%) and a low score (i.e., removed from training data; worst $50$\%) in our $\Sours$.
We confirmed there is almost no difference in diversity between the top and the worst ones in terms of pairs, their utterances, and their responses, respectively.

% ====================================
\subsection{Automatic Evaluation Results for Generated Responses} 
\label{a:generated response with metrics}

Table~\ref{tab:autmatic_evaluations} shows automatic evaluation results for generated responses in our experiment (Section~\ref{sec:case study}).
To calculate these scores, we used publicly available tools.%
    \footnote{\url{https://github.com/moses-smt/mosesdecoder}}$^,$%
    \footnote{\url{https://github.com/Maluuba/nlg-eval}}

%%%%%%%%%%%%%%%%%%%%%%%%%%%%%%%%%%%%
\section{Experimental Results on Japanese} 
\label{a:japanese}
%%%%%%%%%%%%%%%%%%%%%%%%%%%%%%%%%%%%

% ====================================
\subsection{Distributions between Human and Automatic Scoring} 
\label{a:ja-correlation_dist}

\begin{figure*}[]
    \small
    \centering
    \begin{tabular}{c}
        \begin{minipage}{0.33\hsize}
            \centering
            \includegraphics[trim = 0 3mm 0 0, width=0.97\columnwidth]{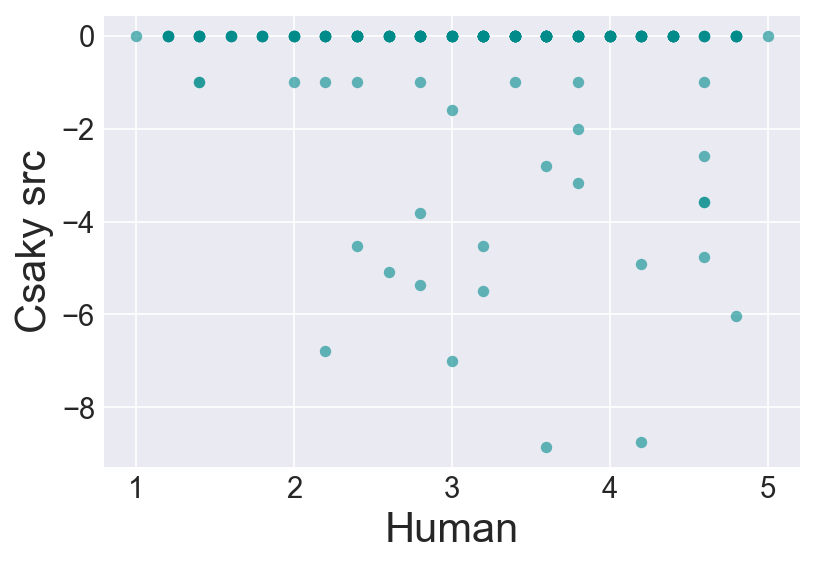}
            \hspace{1.6cm} (a) \citet{csaky2019acl:filtering} SRC
        \end{minipage}
        \begin{minipage}{0.33\hsize}
            \centering
            \includegraphics[trim = 0 3mm 0 0, width=0.97\columnwidth]{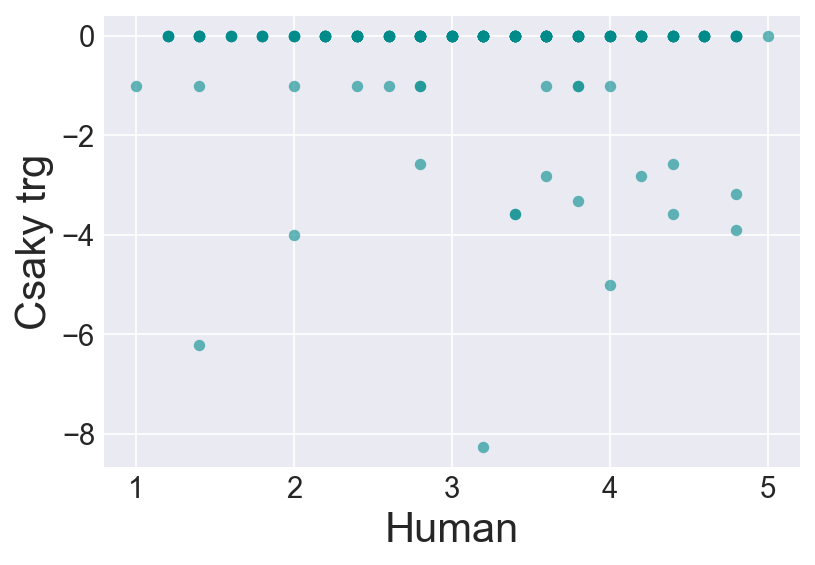}
            \hspace{1.6cm} (b) \citet{csaky2019acl:filtering} TRG
        \end{minipage}
        \begin{minipage}{0.33\hsize}
            \centering
            \includegraphics[trim = 0 3mm 0 0, width=0.97\columnwidth]{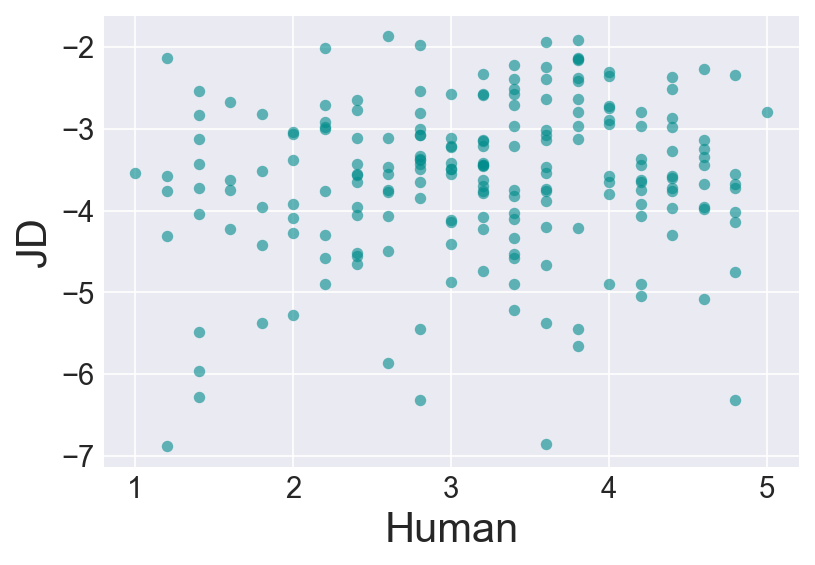}
            \hspace{1.6cm} (c) \citet{junczys-dowmunt2018wmt:filtering}
        \end{minipage}
        \\ \\ \\
        \begin{minipage}{0.33\hsize}
            \centering
            \includegraphics[trim = 0 3mm 0 0, width=0.97\columnwidth]{ja-correlation-human-pairscore-ours.png}
            \hspace{1.6cm}  (d) \, Ours $\Sours$
        \end{minipage}
        \begin{minipage}{0.33\hsize}
            \centering
            \includegraphics[trim = 0 3mm 0 0, width=0.97\columnwidth]{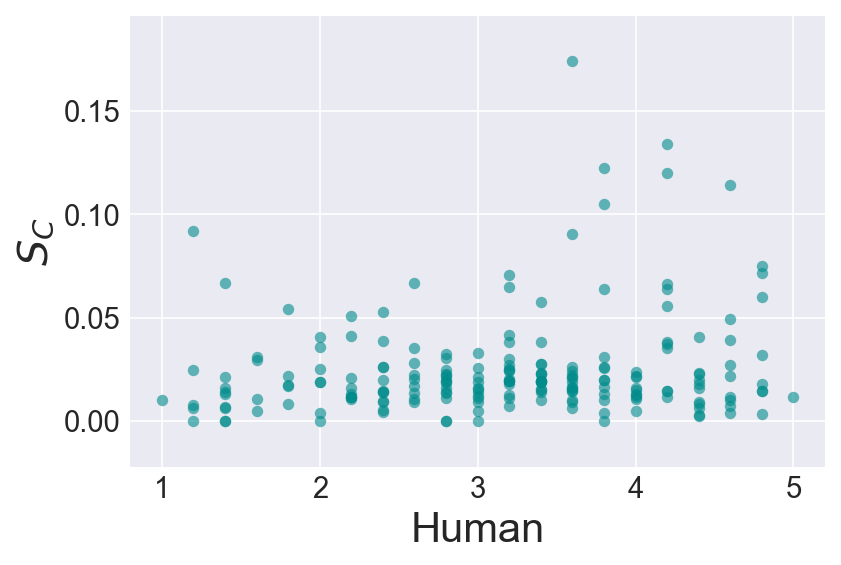}
            \hspace{1.6cm}  (e) \, Ours $\Sframe$
        \end{minipage}
        \begin{minipage}{0.33\hsize}
            \centering
            \includegraphics[trim = 0 3mm 0 0, width=0.97\columnwidth]{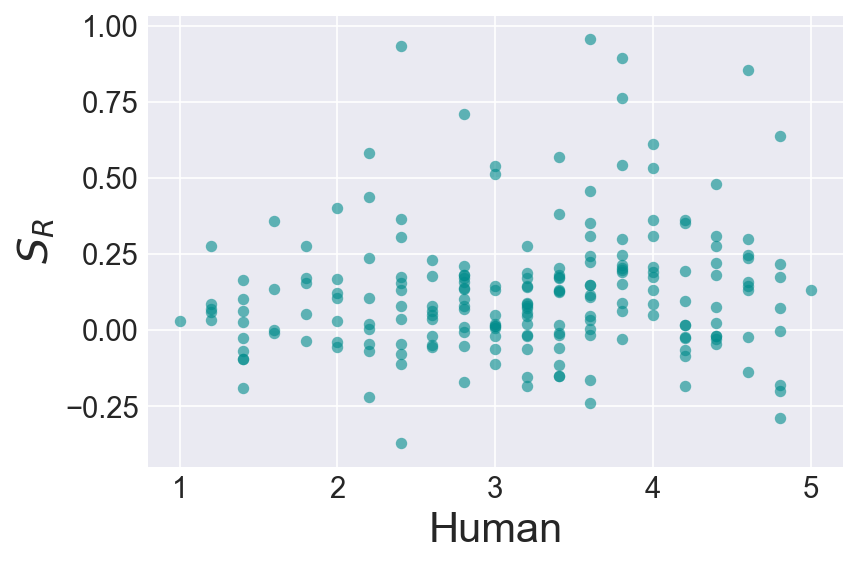}
            \hspace{1.6cm}  (f) \, Ours $\Scontent$
        \end{minipage}
    \end{tabular}
    \caption{Distributions between human scores and automatically computed scores by each method (Japanese).}
    \label{fig:ja_correlation-human-all}
\end{figure*}

Figure~\ref{fig:ja_correlation-human-all} shows that, for all the models including ablations, the distributions between human scores and automatically computed scores.

% =====================================
\subsection{The Distributions of Proposed Scores} 

\begin{figure*}[]
    \centering
    \begin{tabular}{c}
        \begin{minipage}{0.33\hsize}
            \centering
            \includegraphics[width=0.95\columnwidth]{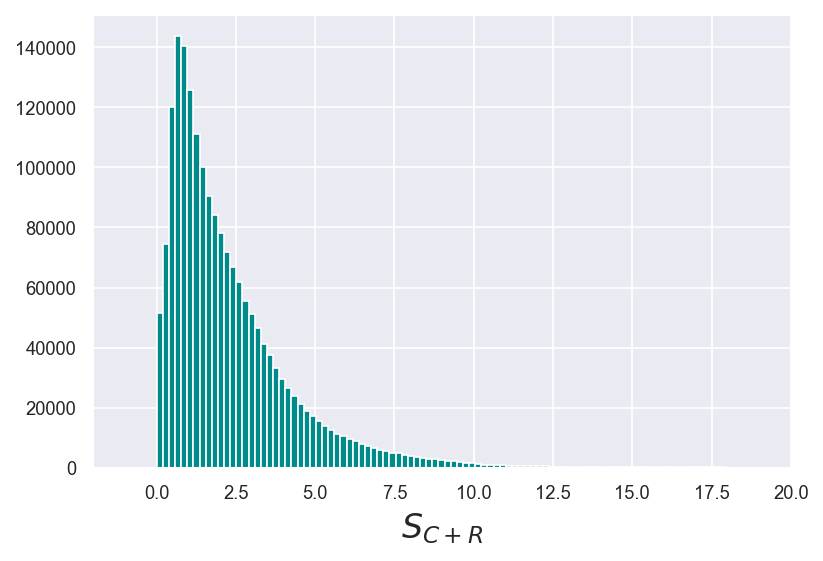}
            \hspace{1.6cm} (a) $\Sours (x,y)$
            \label{fig:s_ours ja}
        \end{minipage}
        \begin{minipage}{0.33\hsize}
            \centering
            \includegraphics[width=0.95\columnwidth]{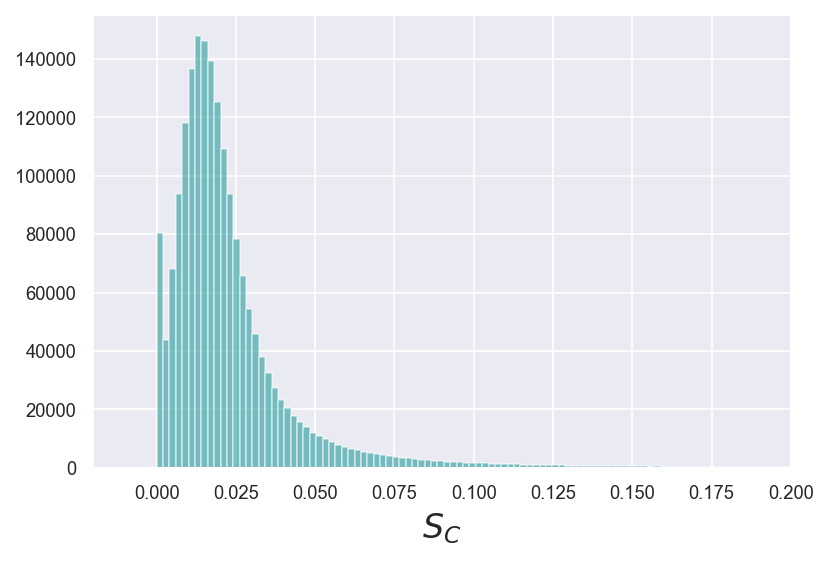}
            \hspace{1.6cm} (b) $\Sframe (x,y)$
            \label{fig:s_frame ja}
        \end{minipage}
        \begin{minipage}{0.33\hsize}
            \centering
            \includegraphics[width=0.95\columnwidth]{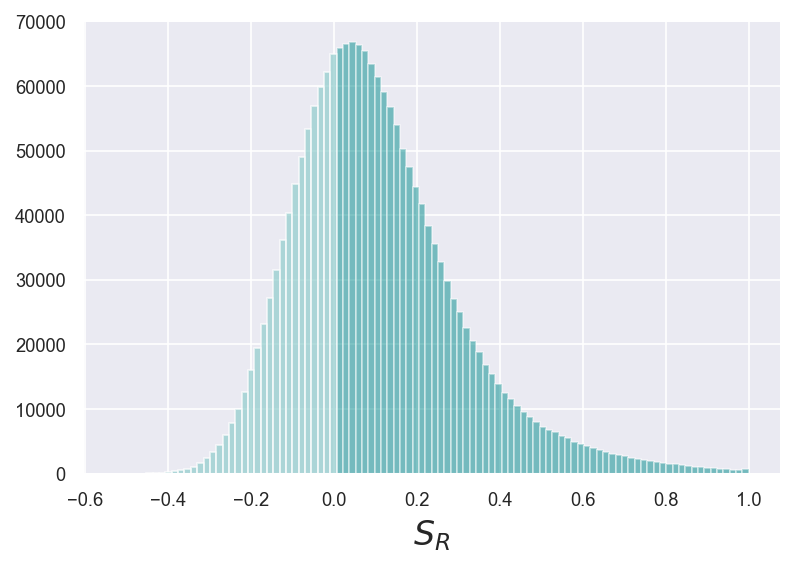}
            \hspace{1.6cm} (c) $\Scontent (x,y)$
            \label{fig:s_content ja}
        \end{minipage}
    \end{tabular}
    \caption{Score distributions of our $\Sframe$, $\Scontent$, $\Sours$ across training data (Japanese).}
    \label{fig:scores ja}
\end{figure*}

Figure~\ref{fig:scores ja} shows that the score distributions of our $\Sframe$, $\Scontent$, $\Sours$ across our training data.
Note that, for computing $\Sours$, $\Scontent$ that less than $0$ (pale-colored part in (c)) are treated as $0$ (Equation~\ref{eq:s_content}).

% ====================================
\subsection{Comparison of Top versus Worst Data} 

Table~\ref{tab:scored-top-vs-worst-ja} shows that the comparison of utterance pairs with a high score and a low score in our $\Sours$.
We confirmed there is almost no difference in diversity between the top and the worst ones in terms of pairs, their utterances, and their responses, respectively.

\begin{table*}[]
    \centering
    \small
    \tabcolsep 4pt
    \begin{tabular}{r lccc lccc lccc}
         \toprule
         \multirow{2}{*}{Scored data} && \multicolumn{3}{c}{Utterance pair} && \multicolumn{3}{c}{Utterance (source-side)} && \multicolumn{3}{c}{Response (target-side)} \\
         \cmidrule{3-5} \cmidrule{7-9} \cmidrule{11-13}
         && len & distinct-1 & distinct-2 && len & distinct-1 & distinct-2 && len & distinct-1 & distinct-2 \\
         \midrule
         Top 50\% (remained)   && 15.00 & 0.023 & 0.245 && 7.51 & 0.041 & 0.366 && 7.49 & 0.041 & 0.367 \\
         Worst 50\% (removed) && 13.79 & 0.025 & 0.260 && 6.90 & 0.045 & 0.378 && 6.90 & 0.044 & 0.374 \\
         \midrule
         Top 90\% (remained)  && 14.62 & 0.015 & 0.176 && 7.31 & 0.028 & 0.286 && 7.31 & 0.028 & 0.285 \\
         Worst 10\% (removed)&& 12.71 & 0.091 & 0.639 && 6.40 & 0.146 & 0.851 && 6.31 & 0.147 & 0.852 \\
         \bottomrule
    \end{tabular}
    \caption{Comparison of the top and the worst utterance pairs in the training data scored by our method (Japanese).}
    \label{tab:scored-top-vs-worst-ja}
\end{table*}

% ==========================================
\subsection{Automatic Evaluation Results for Generated Responses} 

Table~\ref{tab:autmatic_evaluations-ja} shows automatic evaluation results for generated responses in our experiment (Section~\ref{ssec:JapaneseFiltering}).
To calculate these scores, we used publicly available tools.%
    \footnote{For the embedding-based metrics, we used the pre-trained Japanese word embeddings~\cite{grave2018lrec:wordvec157}.}

\begin{table*}[h]
    \centering
    \scriptsize
    \tabcolsep 5pt
    \begin{tabular}{+l^r|^r ^c^c HHH^c^c^c ^c^c^c ^c^c^c}
        \toprule
        Japanese & \# of pairs 
        & len & dist1 & dist2 & B1 & bp & B1bp & B1 & bp & B1bp & MET & ROU & CID & EA & VE & GM \\
        \toprule
        non-filterd & 1,893,477 & 
        5.91 & 268/0.091 & 509/0.207 & 14.0 & 0.79 & 11.1 & 13.4 & 0.79 & 10.6 & 5.44 & 10.98 & 16.27 & 0.723 & 0.438 & 0.585 \\

        \midrule
        \scriptsize{\textbf{Filtered out 10\%:}}  \\
        \citet{csaky2019acl:filtering} SRC & 1,700,000 & 
        5.75 & 295/0.102 & 550/0.231 & 13.9 & 0.77 & 10.6 & 13.2 & 0.76 & 10.0 & 5.09 & 10.79 & 15.03 & 0.711 & 0.430 & 0.575\\
        \citet{csaky2019acl:filtering} TRG & 1,700,000 & 
        7.06 & 336/0.095 & 662/0.219 & 12.0 & 0.97 & 11.6 & 11.6 & 0.96 & 11.1 & 5.75 & 9.91 & 11.23 & 0.730 & 0.434 & 0.581\\
        \citet{junczys-dowmunt2018wmt:filtering} & 1,700,000 & 
        5.31 & 284/0.107 & 516/0.240 & 13.3 & 0.69 & 9.2 & 12.6 & 0.68 & 8.5 & 4.87 & 9.84 & 14.89 & 0.711 & 0.441 & 0.574\\
        Ours $\Sours$ & 1,700,000 & 
        5.68 & 319/0.112 & 582/0.249 & 14.4 & 0.75 & 10.8 & 13.9 & 0.75 & 10.5 & 5.42 & 11.22 & 17.22 & 0.725 & 0.441 & 0.585\\
        Ours $\Sframe$  & 1,700,000 & 
        5.51 & 264/0.096 & 492/0.218 & 14.4 & 0.72 & 10.4 & 13.7 & 0.72 & 9.8 & 5.28 & 10.74 & 15.43 & 0.724 & 0.447 & 0.586\\
        Ours $\Scontent$ & 1,700,000 & 
        5.73 & 296/0.103 & 555/0.234 & 13.2 & 0.76 & 10.1 & 12.5 & 0.76 & 9.5 & 5.20 & 9.85 & 12.82 & 0.719 & 0.441 & 0.579\\

        \midrule
        \scriptsize{\textbf{Filtered out 50\%:}}  \\
        \citet{csaky2019acl:filtering} SRC & 1,000,000 & 
        5.93 & 355/0.120 & 651/0.264 & 11.8 & 0.80 & 9.4 & 11.4 & 0.80 & 9.1 & 5.02 & 9.84 & 13.71 & 0.719 & 0.438 & 0.574\\
        \citet{csaky2019acl:filtering} TRG & 1,000,000 & 
        6.94 & 405/0.117 & 811/0.273 & 12.8 & 0.95 & 12.2 & 12.2 & 0.95 & 11.5 & 5.89 & 10.77 & 13.77 & 0.719 & 0.420 & 0.574\\
        \citet{junczys-dowmunt2018wmt:filtering} & 1,000,000 & 
        5.99 & 421/0.140 & 802/0.321 & 11.7 & 0.81 & 9.4 & 11.2 & 0.79 & 8.9 & 5.02 & 9.25 & 16.41 & 0.706 & 0.421 & 0.561\\
        Ours $\Sours$ & 1,000,000 & 
        5.53 & 405/0.146 & 741/0.327 & 12.9 & 0.73 & 9.4 & 12.4 & 0.72 & 9.0 & 4.95 & 9.20 & 13.58 & 0.707 & 0.428 & 0.565\\
        Ours $\Sframe$ & 1,000,000 & 
        5.48 & 318/0.116 & 599/0.267 & 12.7 & 0.72 & 9.1 & 11.9 & 0.71 & 8.5 & 4.94 & 9.14 & 16.25 & 0.714 & 0.429 & 0.570\\
        Ours $\Scontent$ & 1,000,000 & 
        5.76 & 404/0.140 & 747/0.314 & 13.3 & 0.77 & 10.2 & 12.7 & 0.76 & 9.6 & 5.48 & 10.34 & 18.84 & 0.711 & 0.426 & 0.569\\

        \midrule
        reference &            
        & 7.29 & 750/0.206 & 1446/0.460 & - & - & - & - & - & - & - & - & - & - & - & - \\      
        \bottomrule
    \end{tabular}
    \vskip -3mm
    \caption{Automatic evaluation results for generated responses (Japanese).
    BLEU-1 (B1) and its brief penalty (bp), ROUGE (ROU)$\times 100$, METEOR$\times 100$ (MET), CIDEr$\times 100$ (CID). Embedding-based metrics: Embedding Average Cosine Similarity (EA), Vector Extrema Cosine Similarity (VE), Greedy Matching (GM).}
    \label{tab:autmatic_evaluations-ja}
\end{table*}

\end{document}